\documentclass[journal]{IEEEtran}
\usepackage{amsmath,amssymb,amsthm,bm}
\usepackage{graphicx}
\usepackage{epstopdf}
\usepackage{cite}
\usepackage{algorithm}
\usepackage{algpseudocode}

\usepackage{dsfont}
\usepackage{tikz}
\usetikzlibrary{decorations.pathreplacing,shapes}
\usepackage{pgfplots}
\usepgfplotslibrary{fillbetween}
\usepgfplotslibrary{groupplots}

\usepackage{enumitem}

\usepackage[font=footnotesize]{caption}
\usepackage[font=footnotesize,labelformat=simple]{subcaption}

\usepackage{ifthen}
\newboolean{showcomments}
\setboolean{showcomments}{true}
\usepackage{todonotes}

\IEEEoverridecommandlockouts
 
\theoremstyle{plain}

\theoremstyle{definition}

\theoremstyle{remark}

\DeclareMathOperator{\diag}{diag}

\DeclareMathOperator{\erf}{erf}

\begin{document}

\title{
A Networked Multi-Agent System for Mobile Wireless Infrastructure on Demand  
}
\author{\centering Miguel Calvo-Fullana, Mikhail Gerasimenko, Daniel Mox, Leopoldo Agorio,\\ Mariana del Castillo, Vijay Kumar, Alejandro Ribeiro, and Juan Andr\'es Bazerque
\thanks{
This work was supported by the IoT4Ag Engineering Research Center funded by the National Science Foundation under Award EEC-1941529; and the Agencia Estatal de Investigaci\'on under grant RYC2021-033549-I.

M. Calvo-Fullana is with Universitat Pompeu Fabra, Barcelona, Spain (e-mail: \mbox{miguel.calvo}@upf.edu).

M. Gerasimenko was with Tampere University, Tampere, Finland. He is now with Nokia, Finland. (e-mail: \mbox{mikhail.gerasimenko}@nokia.com).

D. Mox was with University of Pennsylvania, Philadelphia, USA. He is now with Zoox, California, USA. (e-mail: \mbox{dan.c.mox}@gmail.com)

L. Agorio and J. A. Bazerque are with University of Pittsburgh, Pittsburgh, USA (e-mail: \mbox{\{lca31,juanbazerque\}}@pitt.edu).

M. del Castillo is with Universidad de la Rep\'ublica, Montevideo, Uruguay (e-mail: \mbox{mdelcastillo}@fing.edu.uy).

V. Kumar and A. Ribeiro are with University of Pennsylvania, Philadelphia, USA (e-mail: \mbox{\{kumar,aribeiro\}}@seas.upenn.edu).

}}

\maketitle

\begin{abstract}
Despite the prevalence of wireless connectivity in urban areas around the globe, there remain numerous and diverse situations where connectivity is insufficient or unavailable. To address this, we introduce \emph{mobile wireless infrastructure on demand}, a system of UAVs that can be rapidly deployed to establish an ad-hoc wireless network. This network has the capability of reconfiguring itself dynamically to satisfy and maintain the required quality of communication. The system optimizes the positions of the UAVs and the routing of data flows throughout the network to achieve this quality of service (QoS). By these means, task agents using the network simply request a desired QoS, and the system adapts accordingly while allowing them to move freely. We have validated this system both in simulation and in real-world experiments. The results demonstrate that our system effectively offers mobile wireless infrastructure on demand, extending the operational range of task agents and supporting complex mobility patterns, all while ensuring connectivity and being resilient to agent failures.
\end{abstract}


\IEEEpeerreviewmaketitle

\section{Introduction}
\label{sec:Introduction}

Mobile wireless connectivity has become widespread, with access to either Wi-Fi, LTE or 5G channels appearing to be always at reach in most cities around the globe. However, outside urban environments connectivity is still far from ubiquitous. Unfortunately, this disparity is intrinsic, as expanding network infrastructure to remote areas is often unprofitable. Indeed, the low cost of Wi-Fi is offset by its low range. While cellular technologies can reach further, they require significant investment on structures and equipment for cell sites. Even in urban environments, reliable communications are not guaranteed, with occasional high-demand events resulting in temporary network capacity overflows and outages. Dimensioning a system to prevent outages is impractical, as it would imply extraordinary costs to cover for outliers that occur with low probability. 

\begin{figure}
    \centering
    \includegraphics[width=0.95\linewidth]{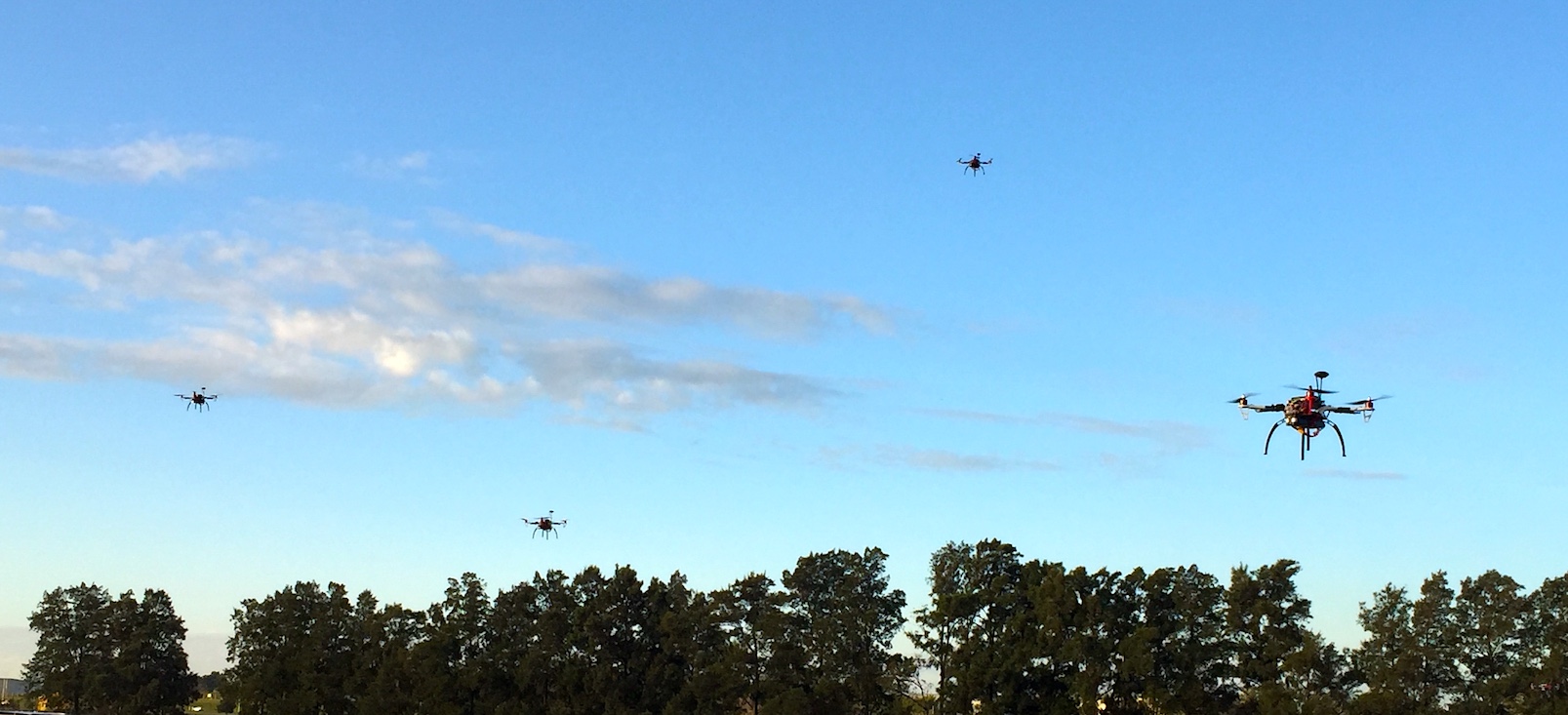}\\    
    \caption{Mobile wireless infrastructure on demand. A team of UAVs acting as network providers continuously reconfigure their positions and communication routes, provisioning task users with their required quality of service in terms of wireless connectivity.}
    \label{fig:swarm}
\end{figure}

\begin{figure}[t]
    \centering
    \includegraphics[scale=1]{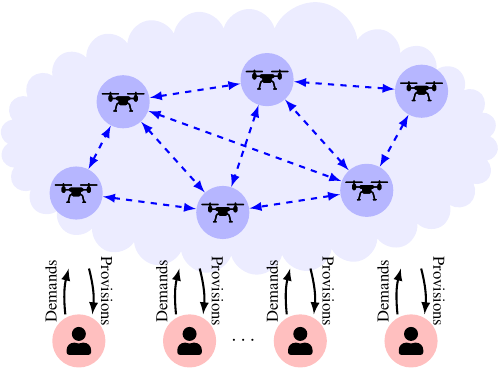}                     
    \caption{Task agents (red) demand a level of connectivity to the MID system, provisioned by the network agents (blue).}
    \label{fig:example_network}
\end{figure}

The concept of mobile infrastructure on demand (MID) has been proposed to address these issues \cite{mox2020mobile}. The idea of MID is to deploy a team of autonomous unmanned aerial vehicles (UAVs) to provide wireless connectivity on demand (Figure  \ref{fig:swarm}). Agents of the MID team reconfigure their positions and communication routes to satisfy quality of service (QoS) requirements of a team of task agents (Figure \ref{fig:example_network}). Task agents can be themselves UAVs or they can be any other type of end user. The distinction between task and MID agents is that the movement of task agents is not controlled, only their communication routes are. 

In this paper, we build a MID system. This is a cyber-physical control system that dynamically reconfigures (physical) positions of the MID agents and (cyber) routing tables of MID and task agents. We considered a series of tests in which task agents move apart, follow undisclosed trajectories, and hold their positions while a MID agent is replaced. Through the evaluation of these tests in simulated scenarios and their corroboration in real experiments, we were able to demonstrate that MID enables the following features:

\begin{description}

\item[(F1)] \textbf{Extension of operational range.} In exploration scenarios the use of a MID team extends the operational range by providing packet relaying capabilities.

\item[(F2)] \textbf{Support for tasks with complex mobility patterns.} In tasks whose execution requires complex mobility patterns MID offers more reliable communication. It does so by continuously adapting positions and routes in response to the movement of the task agents.

\item[(F3)] \textbf{Robust autonomous reconfiguration.} In scenarios in which MID agents fail or exhaust their batteries, other team members reconfigure their positions and routes to limit the effect on end-to-end QoS for the task team.

\end{description}

Features \textbf{(F1)}\textendash\textbf{(F3)} hold significant practical importance across many tasks. One reference application where feature \textbf{(F1)} proves valuable is agricultural monitoring. In this particular scenario, a group of agents moves through a crop field or an orchard, gathering measurements on soil humidity or assessing harvest readiness. These measurements must be relayed back in real-time to a coordination center. Feature \textbf{(F1)} of the MID system allows for larger fields to be covered using the same amount of communication infrastructure, which is now \emph{mobile} infrastructure. An additional example highlighting feature \textbf{(F2)} is a search and rescue mission spanning a large area. In such a scenario, the agents may move in unpredictable ways, and it is crucial for them to maintain communication quality at all times. The reconfiguration capabilities provided by feature \textbf{(F2)} ensure the needed communication QoS. Feature \textbf{(F3)} complements the previous features \textbf{(F1)} and \textbf{(F2)} by providing operational flexibility during task execution. Indeed, autonomous network reconfiguration guarantees the best possible connectivity under adverse conditions, such as the temporary recall or recharge of a UAV during deployment, and prevents the abortion of a mission when coping with agent failures during task execution.

While these examples serve to highlight the system's capabilities in specific applications, our main objective is to offer a flexible solution for providing mobile wireless infrastructure on demand in diverse applications. We are particularly motivated in providing ad-hoc solutions for situations that demand fast deployments during the span of temporary missions. We envision the MID system being utilized in a variety of scenarios, including rescue missions, off-path events, agricultural evaluation, and monitoring of livestock. Additionally, MID can provide wireless service at crowded city gatherings such as sports events, marathons, art festivals, and live concerts. These use cases, among others, serve as a non-exhaustive demonstration of the versatility of our system.

In our modeling abstraction, as depicted in Fig. \ref{fig:example_network}, there is a team of agents carrying out a specific task. Depending on the application at hand, these agents would represent a combination of patrolling UAVs, ground robots collaborating on a mission, livestock with wireless trackers, or even humans using their cell phones or radio equipment. All these scenarios require wireless communication in order to be successful. In this context, our objective is to design a system comprising a second team of agents that provide communication support to the primary team. We argue that UAVs are an appropriate platform for wireless networking as they facilitate fast and flexible deployments and they can reach a proper altitude to provide reliable communication links. Hence, a supplementary team of UAVs will form the system that provides a mobile wireless infrastructure on demand, operating as a set of communication relays that configure an ad-hoc wireless network. While UAVs are ideal for this role, other robotic platforms with different motion dynamics could also be utilized within the same mobile infrastructure system, though they might require distinct motion planning approaches.
 
Compared with standard mobile applications where the positions of the communication base stations are fixed, the ad-hoc network considered here is capable of deploying promptly and being adaptive and robust. The network team can move and track the positions of the task agents in order to maintain and maximize the communication rate they offer. For this optimal design to be possible two technical problems must be solved. First, how to route data across the network in order to maximize throughput for a given spatial arrangement of the task and network teams. Secondly, how to move the network agents to increase the throughput and maintain network connectivity. 

The framework on which we base our system includes an optimization scheme for robust routing introduced in \cite{mox2020mobile}, which takes the positions of the agents in the primary (task) team as parameters. In this paper, we also incorporate Laplacian methods with stochastic communication channels  \cite{fink2011robust,stephan2017concurrent,mox2022learning} to optimize the trajectories of the network agents. A distinctive aspect of our approach is that we define the network team as an autonomous separate entity adapting to the task agents, which do not participate in the path planning. With this concept, the task team can move freely, and our system and control strategy generalizes to provide communication services to a range of applications where task agents could model mobile entities as diverse as robots or livestock.

We further proceed to demonstrate the success of our methods in real-world experiments. For this purpose, we developed an experimental UAV platform shown in Fig. \ref{fig:swarm}, with configurable wireless communication interfaces and sufficient computational power to solve the robust routing and Laplacian optimization in real-time. More crucially, this  platform allows us to demonstrate our system for mobile infrastructure on demand, and showcase the features \textbf{(F1)}\textendash\textbf{(F3)} proposed above, namely, extending the operational range, supporting tasks with complex mobility patterns, and providing a robust reconfigurable network.

\subsection{Related work}

Extensive research effort has been dedicated to addressing the challenge of maintaining interconnectivity among a team of robots \cite{muralidharan2021communication}. Broadly, this research has focused on investigating the integration of communications and control systems in the context of facilitating coordination among autonomous agents. Achieving accurate coordination necessitates careful consideration of both the wireless channel and the communication network. Graph-theoretic approaches have emerged as a common methodology, leveraging graph connectivity as a practical measure to evaluate network strength \cite{spanos2004robust,zavlanos2005controlling,degennaro2006decentralized,zavlanos2007potential,stump2008connectivity,dimarogonas2008decentralized,yang2010decentralized,zavlanos2011graph,ji2007distributed,zavlanos2008distributed}. In these approaches, the edges of the graph represent point-to-point wireless links. These wireless channels can be modeled, in increasing complexity, as binary links \cite{zavlanos2008distributed,ji2007distributed,spanos2005motion,savla2009maintaining}, distance-dependent links \cite{kim2005maximizing,stump2008connectivity,zavlanos2007potential,degennaro2006decentralized,schuresko2009distributed,zavlanos2012network,tekdas2010robotic}, statistical models \cite{mostofi2008communication,mostofi2009characterization,yan2012robotic}, or even direct high-fidelity system-level simulations of the wireless communication system \cite{calvo2021ros,calvo2021communications}. Further works consider, not only connectivity, but also the routing of information through the network \cite{mox2020mobile,williams2014route,varadharajan2020swarm,schack2023robot,stephan2017concurrent,fink2013robust,zavlanos2012network,fink2011communication}. These latter approaches take a more holistic view of the problem. In the case of fixed task agents, \cite{williams2014route} proposes an algorithm that addresses both the placement of network agents to ensure connectivity and their information flow decisions. For dynamic agents, \cite{varadharajan2020swarm} considers maintaining communication flows by dynamically constructing connected graphs through the addition and removal of graph edges, following the approach in \cite{hung2019hierarchical}. Other metrics have also been considered, with \cite{schack2023robot} focusing on placement and flow decisions aimed at minimizing latency. Additionally, approaching information flows as packet routing \cite{zavlanos2012network} adds an additional layer of nuance beyond graph connectivity. For agents attempting to accomplish a task while maintaining communication, \cite{fink2011communication,fink2013robust} develop robust routing schemes. In \cite{mox2020mobile}, task accomplishment is decoupled from communication, introducing the mobile infrastructure on demand system, with robust routing communication and a search-based local planner to control spatial deployment.

Still, the majority of these developments lack extensive experimental validation or remain predominantly theoretical. In this work, we leverage the robust routing principles of \cite{mox2020mobile} and expand on it to support features \textbf{(F1)}\textendash\textbf{(F3)}. This expansion relies in a more intricate graph-based connectivity planner, inspired by the framework developed in \cite{mox2022learning}.

\section{System and Problem Formulation}
\label{sec:problem_formulation}

Consider Fig. \ref{fig:example_network}, illustrating a system with $L=N+M$ agents, where a team of $N$ task agents is deployed in an area of interest to carry out a specific task. These are interconnected by a supplementary team of $M$ network agents that will operate as a set of communication relays. The goal of this section is to find the optimal trajectories and routing strategies for the network team. A prerequisite to answering these questions is to adopt a model for the wireless communication links between agents.  

\subsection{Wireless channel model}
\label{sec:channel_model}

In a free space without obstacles, the wireless transmission of power by an antenna reaches the receptor with a path-loss that depends quadratically on the distance between them  \cite{goldsmith2005wireless}. This quadratic loss of power is inevitable since the wavefront is an expanding surface in space that distributes the transmitted power in different directions. Antenna designs are successful in shaping this surface to direct power towards the intended receptor, with each maker providing corresponding radiation patterns and antenna gains, but the quadratic loss remains.

In more crowded environments, waves are typically obstructed, reflected, and superimposed on their way to the target by interfering objects such as buildings and cars. This further affects how much power reaches the receiving antenna. The path-loss is still modeled as decaying with distance but with an exponent larger than two, whose value is estimated from experimental data. Furthermore, in many cases, communication systems are deployed in environments where the interfering objects move relative to the transmitter-receiver pair. This would be typically the case in an urban environment with passing vehicles. Or even in the case of static objects, such as buildings or trees, which are perceived as moving from the inertial system of moving antennas, for instance, when they are assembled on moving vehicles such as UAVs. In this case, shadowing effects of obstructions and reflections are too complex for a deterministic model, and therefore it is standard to resort to probabilistic models. When adopting a stochastic model, the Shannon information theory identifies the signal-to-noise ratio as the main parameter determining the transmission rate that can be achieved over a communication link.

\begin{figure}[t]
	\centering
   	\includegraphics[scale=1]{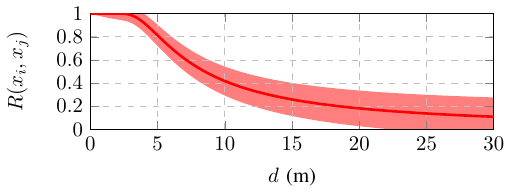}                 
	\caption{Characterization of the expected rate function $R(x_i,x_j)$ and one standard deviation following model equations \eqref{eq:barr} and \eqref{eq:tilder}, where $d=\|x_i-x_j\|$ is the distance between the agents $x_i$ and $x_j$, and $P_{L_0}=-53~\text{dBm}$, $n=2.52$, $P_{N_0}=-70~\text{dBm}$, $a=0.2$ and $b=0.6$.}
	\label{fig:channel_model}
\end{figure}

These stochastic aspects of mobile wireless links are captured in the channel model  \cite{shorey2006mobile} between two antennas at locations $x_i,x_j\in \mathbb R^3$, which is characterized  in Fig. \ref{fig:channel_model}. It relates the transmission power $P_{L_0}$, noise at the receiver $P_{N_0}$, distance $d=||x_i-x_j||$ and path-loss exponent $n$ to the normalized achievable rate $R_{ij}(x_i,x_j) \in [0,1]$. These rates are modeled as stochastic variables with mean $\bar R_{ij}(x_i,x_j)$ and variance $\tilde R_{ij}(x_i,x_j)$ given by  
\begin{equation}
\label{eq:barr}
\bar R_{ij}(x_i,x_j) = \mathrm{erf} \left( \sqrt{\frac{P_{L_0}}{P_{N_0}}\|{x_i-x_j}\|^{-n}} \right) 
\end{equation}
and 
\begin{equation}
\label{eq:tilder}
\tilde R_{ij}(x_i,x_j) = \frac{a \|{x_i-x_j}\|}{b+ \|{x_i-x_j}\|}.
\end{equation}
where $\erf(x)=\frac{1}{\sqrt{\pi}}\int_{-x}^{x}e^{-t^2}dt$ is the Gauss error function and $a$ and $b$ are decay uncertainty parameters. This model is adaptable to diverse hardware by adjusting its parameters to fit empirical measurements \cite{fink2011communication}.

\subsection{Optimal probabilistic routing}
\label{sec:prob_routing}

In practice, the distance between agents is often large enough to cause sufficient path-loss to impede direct communication. Hence, communication occurs over a network of multiple interconnected wireless links distributed across space as in Fig. \ref{fig:example_network}. Information is transmitted by agents, which flows through this network. We consider agents transmitting up to $K$ information flows. These flows consist of data packets, relayed across the network, with different flows representing distinct data with different requirements, e.g., voice and video streaming. The process of relaying data packets is governed by routing variables $\alpha_{ij}^k\in[0,1]$. These variables specify the probability, or fraction of time, that the $i$-th agent sends data to the $j$-th agent for the $k$-th data flow. Further, let the rate $b_i^k(\boldsymbol{\alpha},\mathbf{x})$ be the data injected into the network by the $i$-th agent for the $k$-th flow, with the vector $\boldsymbol{\alpha}\triangleq\{\alpha^k_{ij}\}$ collecting all the routing variables and $\mathbf{x}\triangleq\{x_i\}$ collecting all the agent positions. Then, the difference between data transmitted and received at a node is given by the flow equation
\begin{equation}
\label{eq:flujob}
b_i^k(\boldsymbol{\alpha},\mathbf{x}) = \sum_{j=1}^{L} \alpha_{ij}^k R_{ij}(x_i,x_j)-\sum_{j=1}^{L} \alpha_{ji}^k R_{ji}(x_i,x_j).
\end{equation}
This flow equation \eqref{eq:flujob} indicates that $b_i^k(\boldsymbol{\alpha},\mathbf{x})$ must equal  the aggregated traffic outgoing to the other nodes of the network during the fractions $\alpha^k_{ij}$, after subtracting the incoming traffic from the rest of the network into the node. If the node is a network agent, then it does not inject or absorb data, so that $b_i^k(\boldsymbol{\alpha},\mathbf{x})=0$, and hence all the incoming data is forwarded according to the flow equation. Given \eqref{eq:flujob}, and assuming that the random variables $R_{ij}(x_i,x_j)$ are independent, the mean and variance of the flows in \eqref{eq:flujob} are given by  
\begin{equation}
\label{eq:barb}
\bar b_i^k(\boldsymbol{\alpha},\mathbf{x}) = \sum_{j=1}^{L} \alpha_{ij}^k \bar R_{ij}(x_i,x_j)-\sum_{j=1}^{L} \alpha_{ji}^k \bar R_{ji}(x_i,x_j)
\end{equation}
\begin{equation}
\label{eq:tildeb}
\tilde b_i^k(\boldsymbol{\alpha},\mathbf{x}) \hspace{-0.1ex} = \hspace{-0.1ex} \hspace{-0.3ex} \sum_{j=1}^{L} \left( \alpha_{ij}^k\right)^2 \hspace{-0.5ex}\tilde R_{ij}(x_i,x_j)+\hspace{-0.3ex} \sum_{j=1}^{L} \left( \alpha_{ji}^k\right)^2 \hspace{-0.5ex} \tilde R_{ji}(x_i,x_j).
\end{equation}
Using \eqref{eq:barb} and \eqref{eq:tildeb}, we can establish quality of service (QoS) requirements for the communication network, setting  them as constraints of a routing optimization problem. We consider the case in which the  system must guarantee  minimum rate requirements $b_i^k\geq m_i^k\geq 0$ at each node of the network. For networking agents that are only intended to forward data, we set $m_i^k=0$. On the other hand, each task agent can request a minimum data rate guarantee $m_i^k>0$ from the network. Since the variables involved are stochastic, we must admit a certain probability of outage $1-\epsilon_i^k$. That is, the constraint can only be satisfied probabilistically as follows
\begin{equation}
\label{eq:qos}
\mathbb P \left[ b_i^k (\boldsymbol{\alpha},\mathbf{x}) \geq m_i^k\right] \geq \epsilon_i^k.
\end{equation}
In this case, $1-\epsilon_i^k$ is the probability of outage (the requirement not being satisfied). Thus, the quality of service that a task agent can demand from the network team is the pair $(m_i^k,\epsilon_i^k)$. That is, to provide an average data rate of $m_i^k$ with a confidence level $\epsilon_i^k$. In order to expand expression \eqref{eq:qos}, let us consider achievable rates $R_{ij}(x_i,x_j)$ modeled as Gaussian variables. Then, the following bound holds
\begin{equation}
\label{eq:phiinnverse}
\frac{\bar b_i^k(\boldsymbol{\alpha},\mathbf{x}) - m_i^k}{\sqrt{\tilde b_i^k(\boldsymbol{\alpha},\mathbf{x})}}\geq  \Phi ^{-1} (\epsilon_i^k)
\end{equation}
with $\Phi^{-1}(\cdot)$ being the inverse cumulative Gaussian distribution function. Additionally, we define 
\begin{align}
q_i^k\triangleq \bar b_i^k(\boldsymbol{\alpha},\mathbf{x}) - \Phi ^{-1} (\epsilon_i^k)\sqrt{\tilde b_i^k(\boldsymbol{\alpha},\mathbf{x})}
\end{align}
as the lowest QoS rate, i.e., the statistical minimum rate that needs to be greater or equal to $m_i^k$ for the QoS requirement \eqref{eq:qos} to be satisfied. Furthermore, if the following problem is feasible, then the condition  \eqref{eq:phiinnverse} can be satisfied with an extra margin $s\geq 0$ for all $i$ and $k$, 
\begin{equation}\label{eq:maxs}
\begin{aligned}
	\underset{\boldsymbol{\alpha}\in \mathcal{A}, s \geq 0}{\text{maximize}} & && s \\
	\text{subject to}&
		&& \frac{\bar b_i^k(\boldsymbol{\alpha},\mathbf{x}) - m_i^k-s}{\sqrt{\tilde b_i^k(\boldsymbol{\alpha},\mathbf{x})}}\geq \Phi ^{-1} (\epsilon_i^k) \ \ \ \forall i,k
\end{aligned}
\end{equation}
where the set $\mathcal{A}\triangleq\{\bm\alpha: \alpha_{ij}^k \in[0,1],\ \sum_{i,k} \alpha_{ij}^k\leq 1,\ \sum_{j,k} \alpha_{ij}^k\leq 1\}$ is defined so that the constraint $\bm\alpha\in \mathcal{A}$ makes $\alpha_{ij}^k$ a fraction, and that these fractions can be scheduled without overlapping so that the confluent wireless channels are not overbooked. Solving \eqref{eq:maxs} defines the optimal routing scheme $\boldsymbol{ \alpha}$ satisfying the QoS requirements. Notice that the positions of the UAVs are parameters for this problem so that they are not optimized by solving \eqref{eq:maxs}. Controlling the network agent trajectories is the goal of the ensuing subsection. Problem \eqref{eq:maxs} aims to produce routing tables that maximize the rates across the network. It admits an equivalent reformulation as a convex second-order cone program (SOCP), which ensures that the global optimal solution $\bm\alpha^\star$ and $s^\star$ can be solved efficiently in polynomial time. Appendix \ref{sec:appendixNP} includes the details on how to construct this SOCP from \eqref{eq:maxs}. 

 Implementing the optimal  routing strategy $\bm\alpha^\star$ amounts to agent $i$ transmiting data corresponding to the $k$-th flow to each agent $j$ with probability $(\alpha^k_{ij})^\star$, effectively a fraction of the time. The actual rates achieved when routing according to $\bm\alpha^\star$ are $m_{i}^k+s^\star$, which means that this routing strategy guarantees that all rate requirements $m_i^k$ are satisfied and exceeded with probability $\epsilon_i^k$, and that the surplus of network capabilities is shared equally among agents and flows offering each one an excess of rate equal to $s^\star$.
 
 The rationale behind \eqref{eq:maxs} is that the rate surplus $s^\star$ is maximized by two complementary means. One possibility is to increase the mean rate $\bar b_i(\bm\alpha^\star,\mathbf{x})$ by using the links with higher channel gains $\bar R_{ij}(x_i,x_j)$ more frequently, which amounts to giving higher values to fractions $\alpha_{ij}^k$  corresponding to those links, cf. \eqref{eq:barb}. An alternative strategy to increase the surplus is to reduce the variance $\tilde b_i^k(\bm\alpha,\mathbf{x})$ by selecting the channels with lower variance $\tilde R_{ij}(x_i,x_j)$ and splitting data over multiple nodes, cf. \eqref{eq:tildeb}. Which of these two options is the best one depends on the QoS requirements in terms of rate $m_{i}^k$ and confidence $\epsilon_i^k$, and it is resolved numerically by solving the optimization problem \eqref{eq:maxs}. 

\begin{figure}[t]
   \centering
   \begin{subfigure}{0.49\columnwidth}
   		\centering
   		\includegraphics[scale=1]{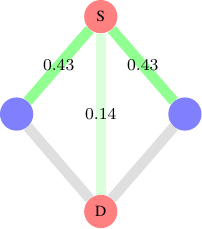}
   		\caption{Low rate margin with high confidence ($m_i^k=0.2$,$\epsilon_i^k=0.95$).}
   		\label{fig:routing1}
   \end{subfigure}
   \begin{subfigure}{0.49\columnwidth}
   		\centering
    	\includegraphics[scale=1]{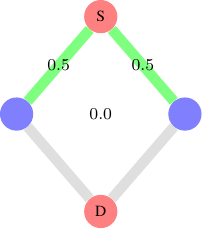}                             
   		\caption{High rate margin with low confidence ($m_i^k=0.5$,$\epsilon_i^k=0.7$).}
   		\label{fig:routing2}
   \end{subfigure}   
   \caption{Illustrative example of the robust routing solution obtained at the source by solving \eqref{eq:maxs}. A low probability of error requires less variance on the transmission rate,  which is achieved by splitting data between the direct link and the relays. On the other hand, a high margin requires all the data to be transmitted to the relays, which present the higher channel gain.}
   \label{fig:routing}
\end{figure}
 
 Fig. \ref{fig:routing} aims to give intuition on the dependence between QoS parameters and routing options in a simple network with a source node  (colored red and marked with an $S$), which sends data to a destination node  (colored red and marked with a $D$), with the option of using their direct link or routing through two relay nodes (colored blue). When high confidence $\epsilon_i^k=0.95$ is required, the variance needs to be brought down, so that the optimal strategy is to split data between the two possible routes. Indeed, given its quadratic form in \eqref{eq:tildeb}, the variance is reduced by dividing the routing variables into smaller fractions. On the other hand, when the confidence margin is relaxed to $\epsilon_i^k=0.7$, but the rate requirement $m_{i}^k$ increases, then the optimal strategy is to route all the data through the relays, which are closer to both the source and the destination, and thus present higher channel gains (cf. Fig. \ref{fig:channel_model}). These two examples also illustrate the robustness to noise of the optimal routing strategy, which takes into account the variance of the channel gains to ensure that the rate requirements are satisfied within the specified confidence margins. It balances mean and variance so that the channels with higher average gains are sacrificed if they  cannot guarantee that the QoS requirements $(m_i^k,\epsilon_i^k)$ are satisfied.

\subsection{Optimal spatial deployment}
\label{sec:spatial_deployment}

Solving the robust routing problem \eqref{eq:maxs} provides us with optimal routing decisions given the positions of the agents. One exceptional aspect of the ad-hoc network of communication devices that we propose is that they are mobile. Hence, the position of network agents can be actively controlled with the goal of increasing data rates. To this end, we consider a network optimization problem over a communication graph. Consider the graph $\mathcal{G}=(\mathcal{V},\mathcal{E})$, where $\mathcal{V}$ is the set of $L$ nodes in the network and $\mathcal{E} \subseteq \mathcal{L} \times \mathcal{L}$ is the set of communication links. The graph is paired with an adjacency matrix  $\mathbf{A}\in \mathbb R^{L\times L}$ that indicate how strong each link is, using the rate model previously introduce in \eqref{eq:barr}, we can define the following entries of the adjacency matrix by
\begin{equation}
\label{eq:graphaij}
A_{ij} \triangleq \bar R_{ij}(x_i,x_j) = \mathrm{erf} \left( \sqrt{\frac{P_{L_0}}{P_{N_0}}\|{x_i-x_j}\|^{-n}} \right).
\end{equation}
The goal of the following optimal controller is to maximize the strength of all links across the network. To do so, we obtain the Laplacian of the graph, given by
\begin{equation}
\mathbf{L} = \mathrm{diag}(\mathbf{A}\mathbf{1})-\mathbf{A}.
\end{equation}
In Appendix \ref{sec:appendixCP} we illustrate how strengthening the connectivity of the graph is equivalent to maximizing the second eigenvalue of the Laplacian \cite{MesbahiEgerstedt2010}, also known as \emph{Fiedler value}. Before proceeding to maximize this  eigenvalue, we consider a linear Taylor approximation $\hat{A}_{ij}$ of the data rates $A_{ij}$ around the current positions $\hat{\mathbf{x}}$ of the agents, which will induce convexity of the ensuing optimization problem. This Taylor approximation is given by
\begin{align}
\hat{A}_{ij} = \hat{R}_{ij}+ \nabla_{x_i} \hat{R}_{ij}^T(x_i-\hat x_i)+ \nabla_{x_j} \hat{R}_{ij}^T (x_j-\hat x_j)
\end{align}
where we have defined $\hat{R}_{ij} \triangleq \bar R_{ij}(\hat{x}_i,\hat{x}_j)$, i.e., the evaluation of the mean rate at the current agent positions. Now we are ready to maximize the Fiedler value, denoted by $\gamma$.
\begin{align}\label{eq:fiedlermax}
	\underset{\mathbf x ,\gamma}{\text{maximize}} & && \gamma \nonumber\\
	\text{subject to}&&& \mathbf{P}^T\hat{\mathbf{L}}\mathbf{P} \succeq \mathbf{I}\gamma \\
			&&& \hat{\mathbf{L}} = \mathrm{diag}(\hat{\mathbf{A}}\mathbf{1})-\hat{\mathbf{A}} \nonumber\\
			&&& \hat{A}_{ij}\hspace{-0.25ex} = \hspace{-0.25ex} \hat{R}_{ij}  \hspace{-0.25ex} +  \hspace{-0.25ex} \nabla_{x_i} \hat{R}_{ij}^T(x_i \hspace{-0.25ex} - \hspace{-0.25ex}\hat x_i)  \hspace{-0.25ex} +  \hspace{-0.25ex} \nabla_{x_j} \hat{R}_{ij}^T (x_j \hspace{-0.25ex}- \hspace{-0.25ex}\hat x_j) \nonumber\\			
			&&& \|\mathbf x-\hat{\mathbf{x}}\|_\infty\leq \Delta \nonumber			
\end{align}
The constraint $\mathbf{P}^T\hat{\mathbf{L}}\mathbf{P} \succeq \mathbf{I}\gamma$, ensures that $\gamma$ is indeed the second eigenvalue of the Laplacian \cite{kim2005maximizing}. For this purpose, the columns of the constant matrix $\mathbf{P}$ must be selected to form a basis of $\mathbf 1^\perp$, that is the complement of the vector of all ones. In addition, the constraint $\|\mathbf x- \hat{\mathbf{x}}\|_\infty\leq \Delta$ is set so that the Taylor approximation remains valid. Appendix \ref{sec:appendixCP} shows that the optimization problem \eqref{eq:fiedlermax} is convex and it can be cast as a semidefinite program (SDP).

All in all, the solution of \eqref{eq:fiedlermax} yields a new set of positions for the network agents that increases the strength of the communication network. An agent designated as the connectivity planner must solve \eqref{eq:fiedlermax} repeatedly, and publish the solution to the network agents, so that they travel across space tracking the sequence of optimal locations.  

\subsection{Network and connectivity planner architecture}\label{sec:architecture}

\begin{figure}[t]
    \centering
    \includegraphics[scale=1]{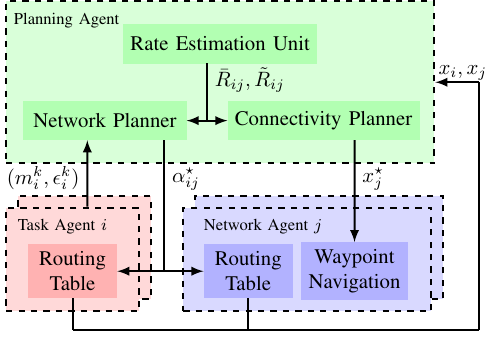}
    \caption{The high-level control loop uses the current positions of network and task agents to estimate the channel rates. Then the network planner selects the optimal routing strategy according to these rates, and the connectivity planner provides trajectories of the network agents to ensure a cohesive communication network.}
    \label{fig:architecture}
    \vspace{-1ex}
\end{figure}

In order to implement the high-level control of the mobile wireless infrastructure on demand a network agent is designated as the planning agent, wherein two planning functionalities coexist (Fig. \ref{fig:architecture}). The network planner implements the optimal probabilistic routing (Section \ref{sec:prob_routing}) and the connectivity planner implements the optimal spatial deployment (Section \ref{sec:spatial_deployment}). These planning functionalities are supported by a rate estimation module (Section \ref{sec:channel_model}).

First, the positions are collected for all network and task agents, i.e. $\{x_i\}_{i=1}^{N}$ and $\{x_j\}_{i=1}^{M}$. These are fed to the \texttt{Rate Estimation Unit}, which computes the mean and variance of channel rates between each pair of agents according to the rate model \eqref{eq:barr} and \eqref{eq:tilder}. These means and variances are inputs for the \texttt{Network Planner}. The \texttt{Network Planner} also takes the QoS requirements $(m_i^k$,$\epsilon_i^k)$ as prescribed configurable parameters, and uses them to specify  problem \eqref{eq:maxs} with variables $\bm \alpha$ and $s$. In particular, $\bm\alpha$ enters the constraint in \eqref{eq:maxs} via \eqref{eq:barb} and \eqref{eq:tildeb}. The equivalent convex SOCP reformulation detailed in Appendix \ref{sec:appendixNP} is solved and the resulting $\bm\alpha^\star$ are shared with all network and task agents. Then each $i$-th agent adjusts its routing table to send data to its $j$-th pair during a fraction $\alpha_{ij}^k$ of the time interval between two consecutive executions of the SOCP solver. Notice that task agents have to be capable of processing these parameters and adapting their routing tables accordingly. 

In parallel, the \texttt{Connectivity Planner} determines the next position of each agent in the network team. From the \texttt{Rate Estimation Unit}, it receives the Taylor approximation of the data rates and uses them together with the set of all current agent positions as parameters for solving the equivalent SDP formulation of \eqref{eq:fiedlermax} shown in Appendix \ref{sec:appendixCP}. The solution $\{x^\star_j\}_{i=1}^{M}$ is published to all network agents, which use them as waypoints to which to navigate. Next, the actual positions are measured and communicated to the planning agent starting a new iteration of the loop in Fig \ref{fig:architecture}. A pseudo-code for the overall planning mechanism is given in Algorithm \ref{alg:NCP}. 

\section{Implementation and Experimental Platform}
\label{sec:implementation}

\begin{algorithm}[t]
\caption{\emph{Network and Connectivity Planner}}\label{alg:NCP}
\begin{algorithmic}
\State Task agent demands QoS $(m_i^k,\epsilon_i^k)$
\While{\texttt{active}}
\State Collect $\{x_i\}_{i=1}^{N}$, $\{x_j\}_{i=1}^{M}$ from task and network agents 
\State \texttt{Rate Estimation Unit}
\State \quad Compute $\bar R_{ij}$, $\tilde R_{ij}$, $\hat{R}_{ij}$, $\nabla_{x_i} \hat{R}_{ij}$, $\nabla_{x_j} \hat R_{ij}$
\State \texttt{Network Planner} 
\State \quad Solve SOCP reformulation of \eqref{eq:maxs} as Algorithm \ref{alg:SOCPplanner}
\State \quad  Distribute solution $\bm \alpha^\star$ to all agents 
\State \texttt{Connectivity Planner}
\State \quad Solve SDP reformulation of \eqref{eq:fiedlermax} as Algorithm \ref{alg:SDPplanner}
\State \quad Distribute solution  $\{x^{\star}_j\}_{i=1}^{M}$ to all network agents 
\EndWhile
\end{algorithmic}
\end{algorithm} 

A main goal of this work is to develop a \emph{system} for mobile wireless infrastructure on demand. To this end, the functionalities presented in the previous section were implemented in ROS and an experimental UAV platform was built to provide each agent with sufficient onboard computational capabilities to solve the optimal network and connectivity planning problems. As described in the previous section, the agent acting as a planning agent must solve \eqref{eq:fiedlermax} in real time to command the MID team to a new set of points in space. It must also solve \eqref{eq:maxs} in order to choose the routing strategy dynamically. Both  \eqref{eq:maxs} and \eqref{eq:fiedlermax} are recast as standard convex programs, and then solved by the convex optimization library \texttt{cvxpy} for Python. The code used to solve \eqref{eq:maxs} and    \eqref{eq:fiedlermax} using \texttt{cvxpy} is wrapped in a planner ROS node that publishes the optimal locations $\mathbf x$ and routes $\bm\alpha$ sequentially.

\subsection{Hardware components}
\begin{figure}[t]
   \centering
   \begin{subfigure}{0.49\columnwidth}
   		\centering
   		\includegraphics[scale=0.22]{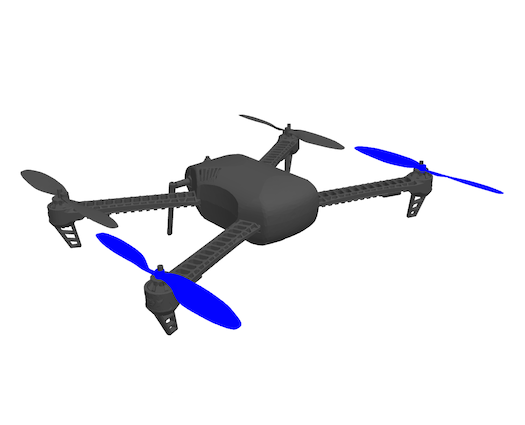}
   		\caption{Simulation platform.}
   \end{subfigure}
   \begin{subfigure}{0.49\columnwidth}
   		\centering
    	\includegraphics[scale=0.22]{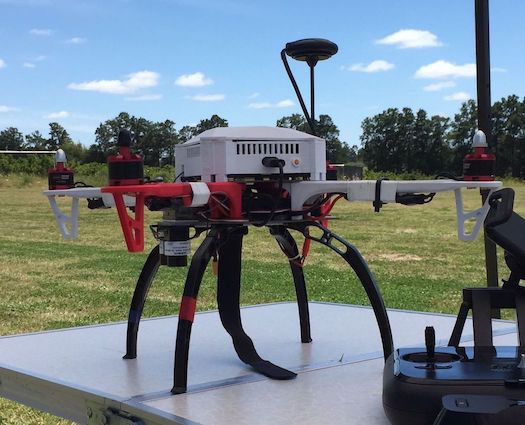}                              
   		\caption{Experimental platform.}
   \end{subfigure}\\[2ex]
   \begin{subfigure}{1\columnwidth}
   		\centering
    	\includegraphics[scale=1]{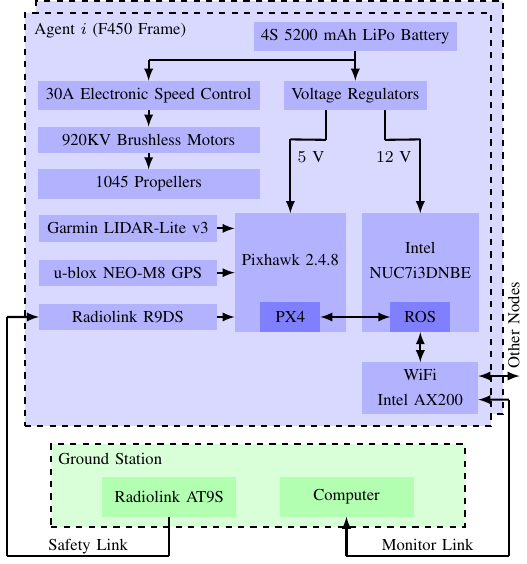}
   		\caption{Hardware components of the experimental platform.}
   \end{subfigure}   
   \caption{Platforms used for simulation and for experimental deployments. A UAV model in Gazebo is used in simulation, while a custom-built UAV platform is used in experiments. The experimental UAV has a low-level flight controller and a high-level processor to optimize the trajectories and routing.}
   \label{fig:sim_platforms}
\end{figure}

For this to be possible, the hardware to be used in experiments must have sufficient compute to be able to support real-time execution of \texttt{cvxpy} over ROS. Most commercial UAV platforms only possess an onboard flight controller with insufficient computation power for our needs. In the absence of an off-the-shelf commercial UAV with these capabilities, we built an experimental quadrotor with both a flight controller and a relatively powerful onboard computer. The choice of a quadrotor, as opposed to other forms of UAVs, has been guided by their availability, flexibility, and simplicity. The main hardware components of this prototype are highlighted in Fig. \ref{fig:sim_platforms}.

A small Intel NUC board is used for this purpose. An Intel 2.4 GHz i3-7100U CPU is mounted on a NUC7i3DNBE motherboard, which is then wired via USB to a Pixhawk Pix 2.4.8 flight controller running the PX4 flight stack. The flight controller receives positioning directives from the NUC via the MAVLink protocol over a USB connection. This low-level flight controller is commanded by MAVROS. The Pixhawk is also provided with signals from a u-blox NEO-M8 GPS receiver, which is taken as a global input to define the position of each agent relative to the common reference of the team. In addition, a downward-facing Garmin LIDAR-Lite v3 is used to provide Pixhawk with a complementary measure of altitude, assisting the PX4 controller at take-off and landing. The two processors and peripherals are mounted on a standard F450 quadcopter frame along with four 920KV brushless motors attached to four 1045 propellers, each one connected to the Pixhawk via a dedicated 30A ESC controller. The F450 structure together with these components is lifted with the power of a lithium polymer battery of four cells and 5200 mAh, which feeds power to the rotors directly, and both to the Pixhawk controller and the onboard computer motherboard, through voltage regulators. 

While the UAVs are designed to fly autonomously, each unit can be connected wirelessly to a manual remote controller (RC) on the ground for safety purposes. For this purpose, a R9DS receiver is wired to the Pixhawk controller onboard and communicates with the RC using spread-spectrum in the band of $2.4$~GHz. Likewise, each UAV has the capability to take flight directives and report flight statistics to a computer on the ground via Wi-Fi. This data can be exchanged with the onboard computer as ROS topics or parameters, or bypassed to the Pixhawk controller via the same MAVLink connection described before.

\subsection{Networking, routing and coordination}

Each UAV agent is assigned a unique IP address within the communication network. The high-level communication among UAVs in the swarm is achieved via IP over an ad-hoc wireless network, which is composed of links between Intel AX200 transceiver cards. These cards are mounted on each onboard computer motherboard, and implement the IEEE 802.11ax wireless standard, radiating power through a pair of dual-band $2.4 - 5.0$~GHz patch Molex 6E antennas. The AX200 cards are configured in IBBS ad-hoc mode operating at $2.4$~GHz, which enables a decentralized network as the one described in Section \ref{sec:problem_formulation}. 

The solution of the robust routing network planner is a probabilistic choice of route. In effect, it indicates the probability with which an agent should use a specific communication link. Equivalently, this is the fraction of time that a link should be used between planning loops. Effectively, the optimal routing strategy obtained by solving \eqref{eq:maxs} is implemented by subsequent modification of the routing tables in the Ubuntu 18.04 (Bionic Beaver) operating system running onboard the NUC. These routing table modifications are effected by the \texttt{ip}, \texttt{ifconfig} and \texttt{route} commands available in the standard set of network tools available in unix-based kernels. An optimal fraction $\alpha_{ij}^k \leq 1$ resulting from solving the optimization procedure is effected in practice by a sequence of probabilistic updates. Specifically, every half a second a Bernoulli variable with probability $\alpha_{ij}^k$ is flipped for each $i,\ j$, and $k$, and if the result is positive the IP of agent $j$ is set as an entry of the routing table of agent $i$.

A key step towards a decentralized multi-agent implementation is to have ROS running in multi-master mode. Setting up this mode also prevents ROS from flooding the wireless network by keeping the ROS topics communicated among agents to a minimum. With each UAV implementing a ROS master, only the topics that report positions and routing fractions are shared among UAVs. The complementary ones needed for each master to run are communicated to the internal localhost IP and do not reach the wireless network, leaving more room for user data. This is an issue of practical nature in the ROS framework. In the future, building a MID system on top of the ROS2 framework, with its Data Distribution Service implementation, would resolve this issue.

Although in our current configuration, both the network planner and connectivity planner run on a designated UAV, and their outputs are broadcast to the other agents via Wi-Fi, all UAVs are built with the same computational capabilities, so that they could run local versions of these planners, which would produce the same outputs, but in a decentralized manner. In both cases, for these mechanisms to work,  agents need to be in a common reference system, this is a practical issue that is addressed in Appendix \ref{sec:appendixCommonReference}.

\section{Simulations}
\label{sec:simulations}

We proceed now to perform a verification of the proposed mobile infrastructure method. To this end, we conduct numerical simulations of the system described in Section \ref{sec:problem_formulation}. The system is implemented in ROS \cite{ros} and simulations are performed with the aid of Gazebo \cite{gazebo}. The hardware used for simulations is comparable to the compute assembled on the experimental platforms. Network connectivity is simulated according to the channel model shown in Fig. \ref{fig:channel_model} and described by equations \eqref{eq:barr}$-$\eqref{eq:tilder}. We consider three different scenarios, each one highlighting a different feature of the mobile wireless infrastructure on demand system. 

\begin{description}
\item[(F1)] Extension of operational range.
\item[(F2)] Support for tasks with complex mobility patterns.
\item[(F3)] Robust autonomous reconfiguration.
\end{description}

All scenarios consider the following premise, a set of task agents is deployed in the environment, performing a specific task. These agents are supported by the MID system. The mobile infrastructure team executes our proposed algorithms and enables the successful accomplishment of the task agents' mission by supporting their network connectivity. For comparison purposes, we also simulate the deployment of static network teams, whose placement is predefined. The fixed system uses the optimal robust routing as described in Section \ref{sec:problem_formulation}, but does not reposition itself. This comparison baseline corresponds to the usual solution of spreading relays in the environment, which has to be done beforehand and are not repositioned dynamically.

\subsection{Extending operational range}
\label{sec:sim1_extending}

\begin{figure}[t!]
   \centering
   \includegraphics[scale=1]{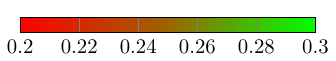}
   \begin{subfigure}{0.49\columnwidth}
   		\centering
   		\includegraphics[scale=1]{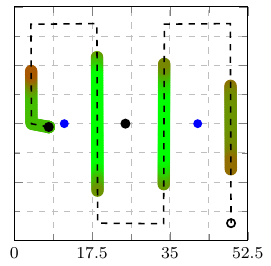}
   		\caption{Rate ($\bar{b}_i^k$) for fixed infrastructure.}
   		\label{fig:sim1_fixed}
   \end{subfigure}
   \begin{subfigure}{0.49\columnwidth}
   		\centering
    	\includegraphics[scale=1]{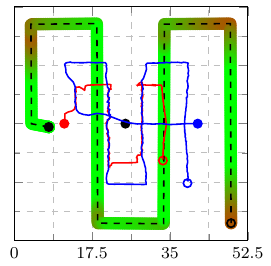}
   		\caption{Rate ($\bar{b}_i^k$) for mobile infrastructure.}
   		\label{fig:sim1_mobile}
   \end{subfigure}\\[2ex]
   \begin{subfigure}{1\columnwidth}
   		\centering
    	\includegraphics[scale=1]{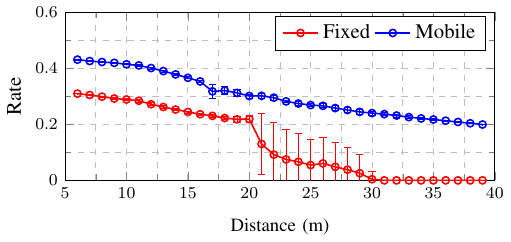} 
   		\caption{Lowest QoS rate ($q_i^k$) with respect to the distance between task agents.}
   		\label{fig:sim1_distance}
   \end{subfigure}   
   \caption{Extending operational range. A task agent performs a square-wave trajectory covering an area of around $50$~m $\times$  $50$~m. This agent must relay information back to another stationary task agent placed in the center of the map. A bidirectional flow of traffic with QoS of $(m_i^k,\epsilon_i^k)=(0.2,0.7)$ is requested. Black dots represent task agents, while red and blue dots represent network agents. Starting positions are shown by a solid dot, while hollow dots correspond to end positions. The lines connecting these dots show the trajectories taken by the agents during the simulation. The rate attained by the task agents experiences a sudden drop when they are $20$~m apart and using fixed infrastructure. In contrast, there is a smooth decay when using the MID system, in which case the range extends up to $40$~m.}
   \label{fig:sim1}
\end{figure}

The first evaluation scenario that we consider is that of operational range extension. The goal here is to demonstrate the capability of the mobile infrastructure to reposition itself and thus cover more area than a predefined fixed deployment. By these means, the MID system can provide a larger operative range with the same number of network agents when compared to a fixed network deployment. Practical applications of this feature match area coverage tasks or data collection situations, in which data must be sent back to a center for post-processing. A clear example would be something akin to agricultural monitoring in which an agent (or a group of agents) moves in a predetermined path through a field, acquiring measurements that must be relayed back in real-time to a coordination center.

For purposes of this simulation, let us consider the scenario illustrated in Fig. \ref{fig:sim1}. In this example, a task agent performs a square-wave trajectory covering the $50$ m $\times$ $50$ m area. Another task agent is placed in the center of the environment and is in charge of collecting information. A bidirectional flow of traffic with QoS given by $m_i^k=0.2$ with confidence $\epsilon_i^k=0.7$ is requested between them. In the case of the fixed configuration, two network agents are deployed in a reasonably equidistant manner over the x-axis. However, the need of covering a square with only two agents clearly leaves the perpendicular y-axis of the map with lower overall coverage, resulting in severe outages along the edges, see Fig. \ref{fig:sim1_fixed}. The same deployment is used in the case of the mobile infrastructure. In this case, the mobile  system is capable of repositioning itself to support the required communication rate along the entirety of the map, cf. Fig. \ref{fig:sim1_mobile}.

This behavior is also certified by looking at the supported rate with respect to the distance between the moving task agent and the receiving task agent (node in the center). This is shown in  Fig. \ref{fig:sim1_distance}. The fixed deployment has a clear cutoff at around $20$ m, where the rate decreases significantly and its variance increases. The latter is due to the fact that moving the same distance along the y-axis results in worse rates than moving the same distance along the x-axis, as the relays are spaced on the x-axis. In contrast, the mobile infrastructure deployment shows a smooth decrease in rate with respect to the distance between agents, and it always  outperforms the fixed infrastructure at the same distance. Specifically,  when the moving  task agent relies on our MID system, in contrast to the fixed setup, it can double the travel distance  from $20$~m  to $40$~m while keeping  communication to the base at a rate over the required minimum of $m_i^k=0.2$. Hence, the mobile infrastructure effectively extends the functional range of the system when operating with the same number of agents.

\subsection{Supporting complex task behaviors}
\label{sec:sim3_patrol}

\begin{figure}[t]
   \centering
   \begin{subfigure}{0.49\columnwidth}
   		\centering
   		\includegraphics[scale=1]{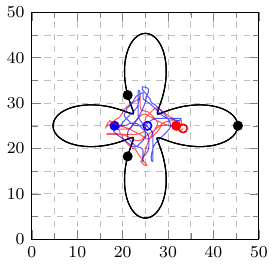}
   		\caption{Agent positions.}
   		\label{fig:sim3_positions}
   \end{subfigure}
   \begin{subfigure}{0.49\columnwidth}
   		\centering
    	\includegraphics[width=0.8\columnwidth]{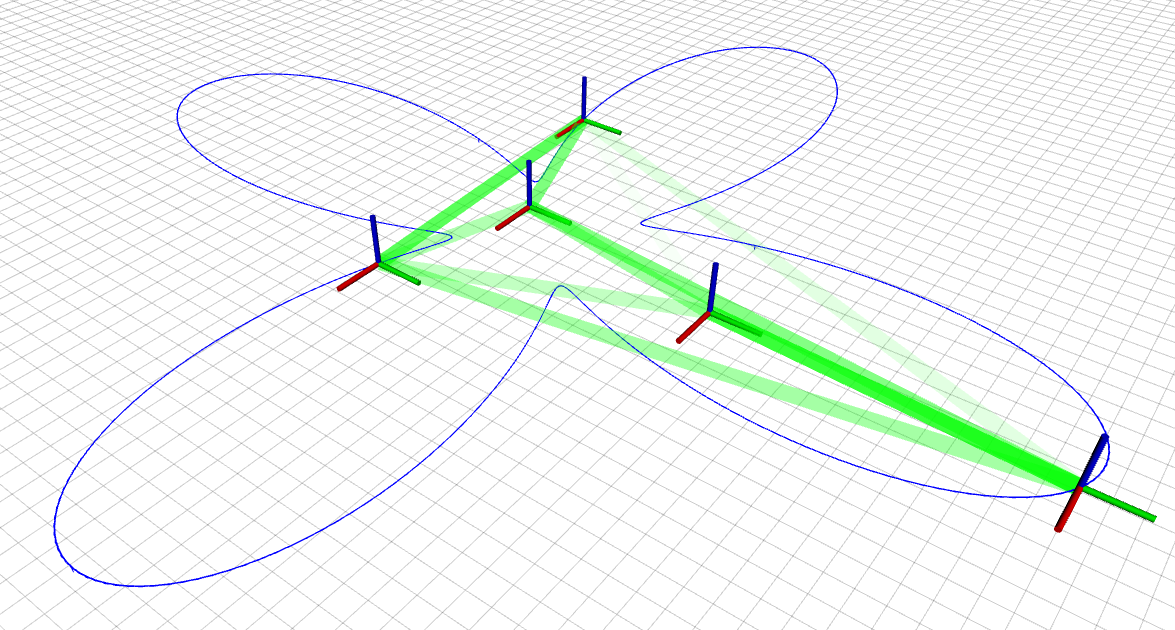}       
   		\caption{Configuration ($t=43.73$ s).}
   		\label{fig:sim3_snap1}   		
    	\vspace{1ex}\includegraphics[width=0.8\columnwidth]{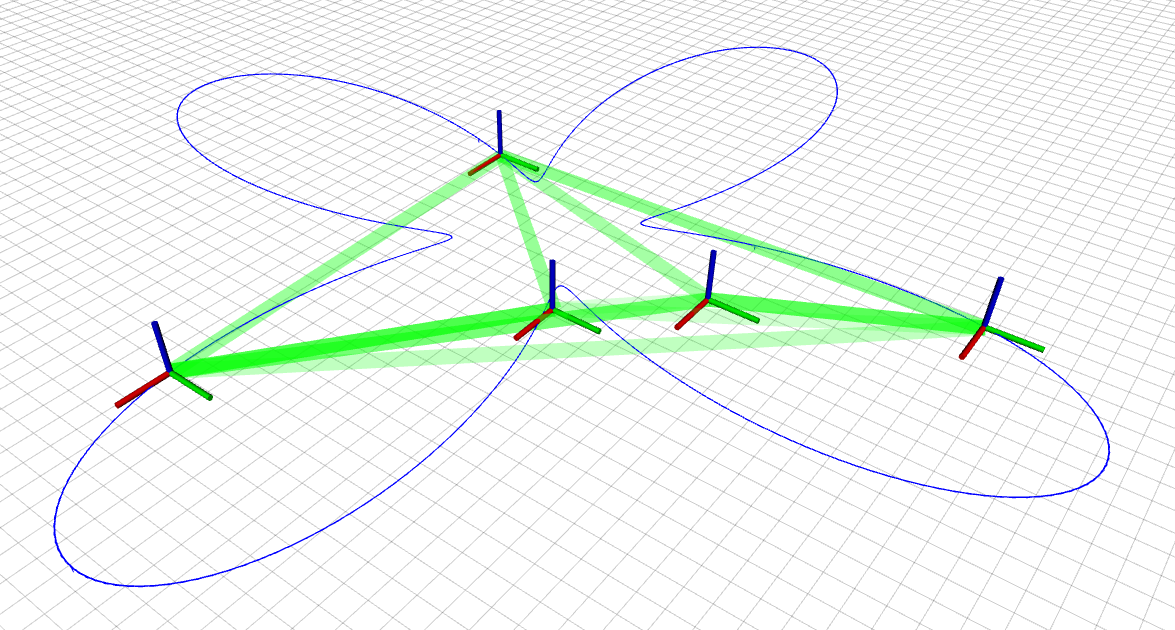}
   		\caption{Configuration ($t=52.46$ s).}   		
   		\label{fig:sim3_snap2}
   \end{subfigure}
   \begin{subfigure}{1\columnwidth}
   		\centering
    	\includegraphics[scale=1]{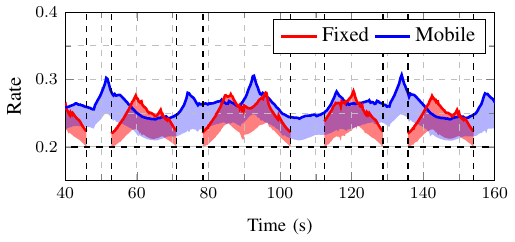} 
   		\caption{Evolution of the mean rate ($\bar{b}_i^k$, solid) and lowest QoS rate ($q_i^k$, shaded).}
   		\label{fig:sim3_rate}
   \end{subfigure}
   \begin{subfigure}{1\columnwidth}
   		\centering
    	\includegraphics[scale=1]{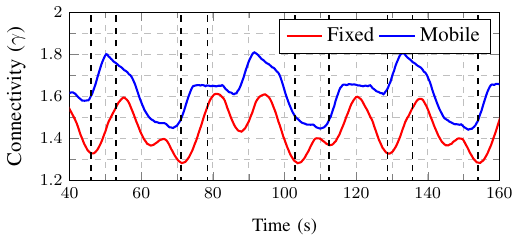} 
   		\caption{Evolution of the algebraic connectivity ($\gamma$).}
   		\label{fig:sim3_connectivity}
   \end{subfigure}\\[2ex] 
   \caption{Supporting complex task behaviors. Three task agents perform the patrol shown by the black line. A QoS of $(m_i^k,\epsilon_i^k)=(0.2,0.9)$ for a circular flow of traffic among task agents is requested. Black dots represent task agents, while red and blue dots represent network agents. Starting positions are shown by a solid dot, while hollow dots correspond to end positions. The lines connecting these dots show the trajectories taken by the agents during the simulation. The probability of routing information among agents is shown by the opacity of the connecting green lines, with more probable routing directions shown as more opaque. In contrast to fixed infrastructure, which suffers from communication outages $28.19\%$ of the time, the MID system guarantees that the QoS requirements are satisfied $100\%$ of the time.}
   \label{fig:sim3}
\end{figure}

In most cases, the behavior of task agents is either unpredictable or follows a complex trajectory. Under these circumstances, deploying fixed infrastructure at optimal locations is not a trivial endeavor. This fact either leads to denser infrastructure deployments than needed, or QoS not being satisfied due to a lack of network agents. Next, we study one such scenario. We consider three task agents conducting a patrol following the four-leafed clover trajectory shown in Fig. \ref{fig:sim3}. Two network agents are employed to support a circular flow of traffic among the task agents. In this case, the node placement of the fixed infrastructure is not trivial, as task agents might go in and out of sync of being close or far from the center of the trajectory. Thus, we place the fixed infrastructure across the x-axis. The exact placement of the two network agents for the fixed infrastructure is shown by the solid red and blue dots in Fig. \ref{fig:sim3_positions}.

The mean rate $\bar b_i^k$ and lower QoS rate $q_i^k$ for agent $i=1$ and flow $k=1$ are shown in Fig. \ref{fig:sim3_rate}. The red and blue shaded bands correspond to rates obtained using the fixed infrastructure and MID system, respectively. They extend between the mean rate $\bar b_i^k(t)$ (solid line) and the lowest QoS rate $q_i^k(t)$ (lowest shaded region). This figure illustrates that while fixed infrastructure suffers from frequent communication outages, mobile infrastructure is capable of reconfiguring dynamically  and supporting the required rate in a relatively stable manner. Due to the use of only two network agents, fixed infrastructure is unable to fully cover the trajectory of the task agents. Specifically, the fixed setup cannot hold the required QoS during recurrent time intervals, corresponding to communication outages $28.19\%$ of the total time span of the simulation. The naive fixed deployment used spreads the network agents over the x-axis, thus when task agents move far in the y-axis, communication outages occur. In contrast, the capability of the mobile infrastructure to dynamically adapt means that two network agents can efficiently cover the whole trajectory. As seen in  Fig. \ref{fig:sim3_positions}, the overall trajectories of the network agents stay relatively close in range to the naive fixed infrastructure locations. However, they rotate and adapt to track the movement of the task agents. 

The movement of the network agents is better analyzed by snapshots of their configuration shown in Fig. \ref{fig:sim3_snap1}\textendash\ref{fig:sim3_snap2}. The mobile infrastructure places network agents inside the triangle connecting the task agents, attempting to stay relatively close to the center of it. When one of the task agents moves far away in one axis while the other two task agents remain close (e.g., Fig. \ref{fig:sim3_snap1}), the network agents elongate their positions maintaining the quality of service required. In this case, fixed infrastructure is incapable of maintaining connectivity and produces a communication outage (see Fig. \ref{fig:sim3_rate}, $t=43.73$ s). Furthermore, routing decisions are also affected by the agents' positions. For instance, at $t=52.46$ s (Fig. \ref{fig:sim3_snap2}), the task agents are at a point in their trajectories such that they are closer to each other, this results in routing decisions that are more spread over with some probability of traffic going directly from task agent to task agent. With the same number of network agents, in contrast to fixed infrastructure which suffers from communication outages $28.19\%$ of the time, the MID system guarantees that the QoS requirements are satisfied $100\%$ of the time. Thus, the MID system adapts to support task agents with complex mobility patterns where fixed infrastructure fails.

Another interesting aspect to study is the relationship between the resulting rate, Fig. \ref{fig:sim3_rate}, and the algebraic connectivity, Fig. \ref{fig:sim3_connectivity}. Recall that the measure of true interest is rate and we use algebraic connectivity as a surrogate to design the connectivity planner \eqref{eq:fiedlermax}. Fig. \ref{fig:sim3_connectivity} shows that, in effect, algebraic connectivity serves as a good proxy measure for the resulting rate of the system.
  
We have further used this simulation scenario to obtain estimates of the time required to execute the network and connectivity planning loop (Algorithm \ref{alg:NCP}). Independent simulations with the same task agent trajectories and QoS demands and network agents ranging from 2 to 14 nodes show average re-planning times of $0.45~\text{s}$ to $5.06~\text{s}$ with standard deviations of $0.019~\text{s}$ and $0.124~\text{s}$, respectively.

\subsection{Robust autonomous reconfiguration}
\label{sec:sim2_replacement}

\begin{figure}[t]
   \centering
   \begin{subfigure}{0.49\columnwidth}
   		\centering
   		\includegraphics[scale=1]{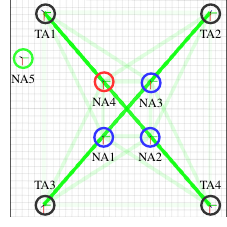}
   		\caption{Initial Configuration.}
   		\label{fig:sim2_config_initial}
   \end{subfigure}
   \begin{subfigure}{0.49\columnwidth}
   		\centering
    	\includegraphics[scale=1]{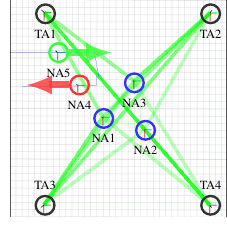}
   		\caption{Exchange Reconfiguration.}
   		\label{fig:sim2_config_mid}
   \end{subfigure}\\[2ex]
   \begin{subfigure}{1\columnwidth}
   		\centering
    	\includegraphics[scale=1]{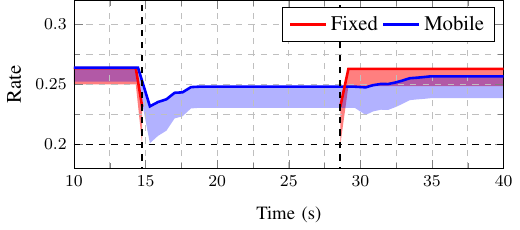} 
   		\caption{Evolution of the mean rate ($\bar{b}_i^k$, solid) and lowest QoS rate ($q_i^k$, shaded).}
   		\label{fig:sim2_rate}
   \end{subfigure}   
   \caption{Robust autonomous reconfiguration. Task agents are the outside square of four agents. The network agents are the inner set of four agents plus the substitute. A flow of traffic with QoS of $(m_i^k,\epsilon_i^k)=(0.2,0.7)$ is requested from node 2 to node 4 and the same requirements for a flow from node 3 to node 1. The MID system prevents a connectivity outage of $12$~s that would result from using a static setup in the event that a networking node is deactivated and needs to be replaced.}
   \label{fig:sim2_config}
\end{figure}

We now consider another simulation case wherein one network agent must be replaced by another. The motivation underlying this scenario is practical. In many cases, fixed infrastructure is deployed to cover a large dense area. Some of the fixed nodes in this infrastructure might need to be replaced due to malfunction, battery limitations, or simply due to the need for continuous operation or network design. Specifically, for purposes of the simulation, in this scenario, task agents are simply stationary. We set a single network agent to be replaced by a different agent as shown in Fig. \ref{fig:sim2_config}. This simulation consists of an outer square of task agents, requesting a flow of traffic among some of them. An inner square of network agents is deployed to support these communication needs. Under this initial configuration, each task agent mainly routes information to its nearest network agent, with alternative routes occurring with low probability.

At a specific point during the simulation, the top left network agent depicted in red deactivates itself from the network configuration and needs to be replaced by another network agent. During this replacement interval, the agents in the mobile infrastructure reconfigure themselves to provide the required connectivity while the new agent arrives at the formation. Figures \ref{fig:sim2_config_initial}\textendash\ref{fig:sim2_config_mid} shows the clear change in the agents' positions. Network agents move slightly to cover the region of the network agent being replaced. The routing decisions are also altered. Information flow is more evenly spread through the network, in contrast to it being routed with high probability to the closest network agent. As the replacement network agent enters the network, the nodes begin to reposition themselves again, reaching the same  initial configuration with four agents as shown in Fig. \ref{fig:sim2_config_initial}. The temporary reconfiguration is necessary to maintain the quality of service requested by the task agents. In contrast to this adaptive solution, removing a single agent from the fixed infrastructure might result in the network being unable to support the required rates. This is the case of this simulation, as it is apparent from the mismatching data rates shown in Fig. \ref{fig:sim2_rate}.

 In the static case, when the network agent deactivates itself to be substituted and goes out of the system, the lowest QoS rate ($q_i^k$, the lower limit of the shaded region) drops under the required threshold, resulting in a connectivity outage. This outage lasts for the complete duration of the replacement procedure ($15$~s to $27$~s in Fig. \ref{fig:sim2_rate}). In contrast, the mobile infrastructure immediately reconfigures itself, recovering the rate and managing to maintain connectivity during the network agent exchange process. When the replacement agent arrives, the network configuration goes back to the same initial configuration shown in Fig. \ref{fig:sim2_config_initial}.

This illustrates the ability of the mobile infrastructure system to provide more than simply extended coverage. Since a fixed infrastructure design needs to place nodes in specific predetermined positions, a failure of a single node can be catastrophic to overall network connectivity unless we explicitly introduce redundancy into the system, with the corresponding increase in the number of required nodes. In contrast, the mobile infrastructure system is robust to system failure by providing a capable reconfiguration mechanism to maintain the requested level of connectivity.

An important observation common to the three scenarios presented in this section  is the fact that even if  these are simulated experiments, they run in real-time on similar hardware to the real-life experimental platform. A consequence of this is that there is a small delay between reconfigurations, as the agents need to solve the optimization problems discussed in Section \ref{sec:problem_formulation}. Specifically, for the simulation in this section, the replanning process runs at around 2.85 Hz. This publishing rate could be increased with better computing power, but  the current case  is representative of what will be observed in experimental deployments. Nonetheless, while faster computers (or avoiding real-time operation) would lead to better rate results due to more adaptive replanning, mobile infrastructure outperforms fixed infrastructure even with the current conditions.

In this section, we have presented three illustrative instances that showcase the capabilities of the MID system. Firstly, we have successfully expanded the operational range, feature \textbf{(F1)}, effectively doubling it for an area coverage task. Secondly, in a task involving intricate mobility patterns, feature \textbf{(F2)}, we have effectively mitigated communication outages from 28.19\% to 0\%, entirely eliminating disruptions in a complex patrol scenario. Thirdly, our last simulation reveals the inherent robustness of the MID system, feature \textbf{(F3)}. In a scenario where an agent failure occurs, the MID system instantaneously reconfigures itself to seamlessly support network requirements, avoiding a 12 second outage, thereby completely avoiding any task disruptions.

\section{Experiments}
\label{sec:experiments}

Complementarily to the simulations in the previous section, we performed experimental demonstrations. These experiments are conducted using the hardware platforms described in Section \ref{sec:implementation}. These UAVs implement the required algorithms and networking capabilities to test our proposed MID system. We configure the UAVs' wireless cards to operate in ad-hoc mode. Hence, the nodes themselves create their own mesh network, which they use for communication. To test the network capabilities we use the \texttt{iperf3} and \texttt{traceroute} tools of the Linux system to measure both the actual throughput and delay. Effectively, these tools do the following: \texttt{iperf3} injects TCP/IP or UDP packets into the network interface of choice and with a given destination, allowing for the measurement of communication throughput; on the other hand, \texttt{traceroute} provides the round trip time of packets between two nodes, to measure the delay experienced in the network.

\begin{figure}[t]
   \centering
   \begin{subfigure}{0.49\columnwidth}
   		\centering
   		\includegraphics[scale=1]{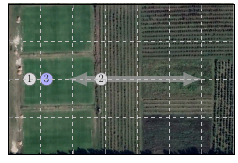}
   		\caption{Location of Experiments.}
   		\label{fig:exp1_snapshots_location}
   \end{subfigure}
   \begin{subfigure}{0.49\columnwidth}
   		\centering
    	\includegraphics[scale=1]{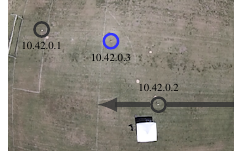}                             
   		\caption{Range Experiment.}
   		\label{fig:exp1_snapshots_experiment}
   \end{subfigure}   
   \caption{Top-down view of the location used for the experimental sessions, where each square is approximately $50$ m $\times$ $50$ m. A snapshot of the range extension experiment being conducted is shown on the right. Task agent $10.42.0.2$ progressively moves away from the stationary task agent $10.42.0.1$, relaying information back to it. A mobile infrastructure agent $10.42.0.3$ redeploys itself dynamically to support this experiment.}
   \label{fig:exp1_snapshots}
\end{figure}

The location used for conducting the experiments is on the outskirts of Montevideo, Uruguay. This area, shown in Fig. \ref{fig:exp1_snapshots_location}, spans several football and agricultural fields. Due to its location, minimal wireless interference exists in this area. Similarly to the simulations in the previous section, we conduct three experiments, each highlighting a feature \textbf{\textbf(F1)}\textendash\textbf{\textbf(F3)} of the MID system.

\subsection{Extending operational range}

\begin{figure}[t]
   \centering
   \begin{subfigure}{0.49\columnwidth}
   		\centering
   		\includegraphics[scale=1]{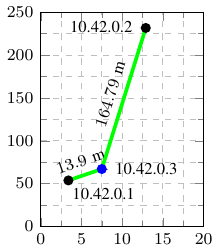}
   		\caption{Fixed infrastructure ($d=178.23$ m).}
   		\label{fig:exp1_position_fixed}
   \end{subfigure}
   \begin{subfigure}{0.49\columnwidth}
   		\centering
    	\includegraphics[scale=1]{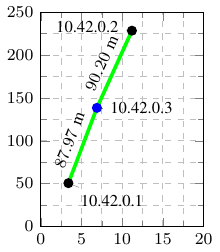}
   		\caption{Mobile infrastructure ($d=178.17$ m).}
   		\label{fig:exp1_position_mobile}
   \end{subfigure}\\[2ex]
   \begin{subfigure}{1\columnwidth}
   		\centering
    	\includegraphics[scale=1]{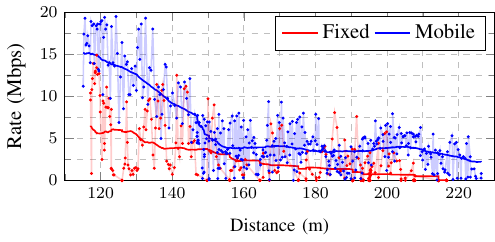} 
   		\caption{Rate with respect to the distance between task agents. Individual rate measurements and a moving average over 25 meters are shown.}
   		\label{fig:exp1_rate}
   \end{subfigure}   
   \caption{Extending operational range. The MID system operates above $2.5$~Mbps at distances of up to $220$~m, while the fixed infrastructure can only support this data rate for up to $160$~m. While the fixed infrastructure is limited by its predetermined placement, the mobile infrastructure places itself approximately in the middle of the two task agents, supporting longer ranges and overall higher rates than the fixed infrastructure.}
   \label{fig:exp1}
\end{figure}

In the first experiment, the MID system extends the operational range of the agents, similar to the previously performed simulation in Section \ref{sec:sim1_extending}. The node placement for this experimental test is shown in Fig. \ref{fig:exp1_snapshots_experiment}. We label nodes by their IP addresses in the ad-hoc network. In this experiment, a stationary task agent (10.42.0.1) is placed at one end of the experiment area. This node takes off but does not move, maintaining a hovering state. Another task agent (10.42.0.2) takes off and moves away from the first agent in a straight line. A network agent is placed in the middle (10.42.0.3) which effectively acts as a relay between the two task agents. We conduct two distinct tests, one with fixed infrastructure placed at $20$~m away from the stationary task agent, in which the network agent takes off but does not reposition itself, simply executing the solution to the robust routing problem. Then, we conduct another test using the mobile infrastructure, in which the network agent repositions itself dynamically using the MID system. 

The results from this experiment are shown in Fig. \ref{fig:exp1}. We measure the communication rate between the task agents and plot it against their distance from each other (Fig. \ref{fig:exp1_rate}). The capability of the MID system to extend the operational range is apparent from Fig. \ref{fig:exp1_rate}, since the rate at any given distance is always lower when using the fixed infrastructure. Due to the limited wireless interference at the location of these experiments, the antenna directivity pattern and the complete line of sight between agents, very long operational ranges can be achieved. Nevertheless, the mobile infrastructure is shown to be capable of operating at around $2.5$~Mbps at distances of up to $220$~m, while the fixed infrastructure can only support this data rate at distances of up to $160$~m.

\begin{figure}[t]
   \centering
   \begin{subfigure}{0.49\columnwidth}
   		\centering
   		\includegraphics[scale=1]{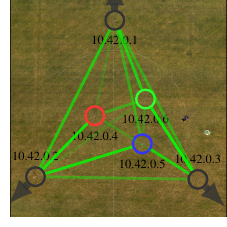}
   		\caption{Configuration ($t=5$ s).}
   		\label{fig:exp3_config_initial}
   \end{subfigure}
   \begin{subfigure}{0.49\columnwidth}
   		\centering
    	\includegraphics[scale=1]{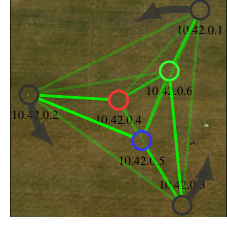}                             
   		\caption{Configuration ($t=145$ s).}
   		\label{fig:exp3_config_extended}
   \end{subfigure}\\[2ex]
   \begin{subfigure}{0.49\columnwidth}
   		\centering
    	\includegraphics[scale=1]{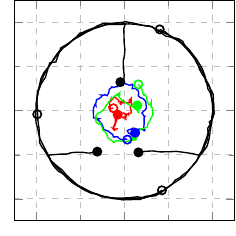}                             
   		\caption{Trajectories. Grid space is $10$ m.}
   		\label{fig:exp3_trajectories}
   \end{subfigure}   
   \begin{subfigure}{0.49\columnwidth}
   		\centering
    	\includegraphics[scale=1]{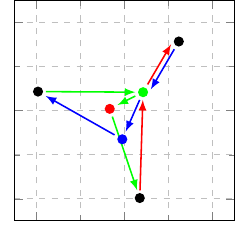}                             
   		\caption{Executed routing ($t=145$ s).}
   		\label{fig:exp3_routing_executed}
   \end{subfigure}         
   \caption{Spatial configurations and routing solutions of the mobile infrastructure during the patrol experiment. Three task agents rotate in a counter-clockwise circle of $40$~m diameter. Three network agents are deployed to provide the required connectivity.}
   \label{fig:exp3_snapshots}
\end{figure}

The optimal placement of the network node at such long distances is intuitive, one should place it in the middle between the task agents. We verify this in Fig. \ref{fig:exp1_position_fixed}\textendash\ref{fig:exp1_position_mobile}. While the relay is incapable of moving when the infrastructure is fixed,  the network agent in the mobile system always keeps itself near the center point between the two task agents. This enables the extended operational range. Since the data rate decreases with respect to distance (See e.g., Fig. \ref{fig:channel_model} and equation \eqref{eq:barr}, for the channel model), this results in overall better data rates.

This experiment verifies that mobile infrastructure is capable of making more efficient use of network agents. Indeed, by dynamically routing and repositioning, further operational ranges can be achieved without the need of additional agents. The MID system also proved itself robust to model inaccuracies. The parameters fitted in Fig. \ref{fig:channel_model} resulted to be too conservative in the experiment site probably due to low noise levels. However, problem \eqref{eq:fiedlermax} remains feasible so that the connectivity planner will always work under model mismatch, and the network planner can protect itself by lowering the parameter $m_i^k$ if needed, as the border case $m_i^k=0$ ensures the feasibility of \eqref{eq:maxs}. 

\subsection{Supporting complex task behaviors}

We proceed with a second experiment consisting of a circular patrol by three task agents. This experiment emulates the complex behavior simulation discussed in Section \ref{sec:sim3_patrol}. For practical reasons, in the following two experiments, we opted to position the antennas on the arms of the UAV frame as opposed to the legs. This deliberate placement significantly diminishes the antenna gain within the plane parallel to the ground. While this outcome would typically be considered undesirable, it enables us to conduct multi-agent experiments within a more confined space. In this experiment, three task agents (10.42.0.1, 10.42.0.2, and 10.42.0.3) and three network agents (10.42.0.4, 10.42.0.5, 10.42.0.6) are deployed utilizing the MID system. As the experiment initiates, the task agents move into a circular patrol of $40$~m diameter, rotating in a counter-clockwise manner. 

\begin{figure}[t]
   \centering
   \begin{subfigure}{1\columnwidth}
   		\centering
   		\includegraphics[scale=1]{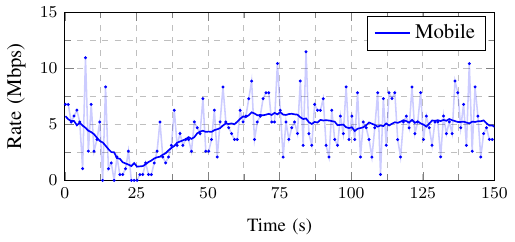}
   		\caption{Average data rate in the network.}
   		\label{fig:exp3_rate}
   \end{subfigure}\\[2ex]
   \begin{subfigure}{1\columnwidth}
   		\centering
    	\includegraphics[scale=1]{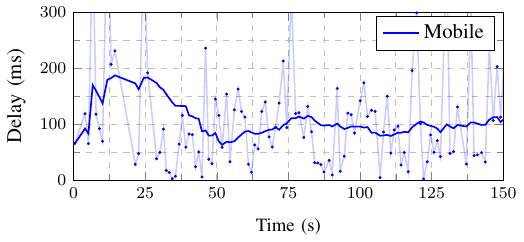}                             
   		\caption{Average delay in the network.}
   		\label{fig:exp3_delay}
   \end{subfigure}   
   \caption{Patrol experiment. Instantaneous rate and delay measurements are collected every second and a moving average over 20 seconds is shown. The average data rate and delay achieved by the MID system stabilizes after $~60$~s and is kept around $5$~Mbps and below $115$~ms.}
   \label{fig:exp3_rate_delay}
\end{figure}

\begin{figure*}
   \centering
   \begin{subfigure}{0.32\linewidth}
   		\centering
   		\includegraphics[scale=1]{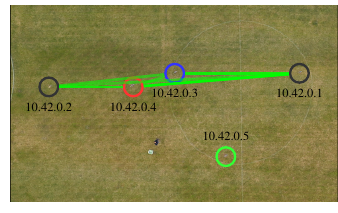}
   		\caption{Initial configuration ($t=3$ s).}
   		\label{fig:exp2_config_initial}
   \end{subfigure}
   \begin{subfigure}{0.32\linewidth}
   		\centering
    	\includegraphics[scale=1]{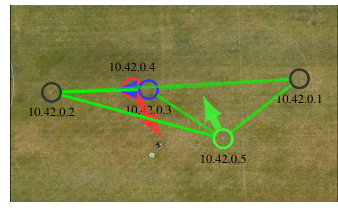}                             
   		\caption{Exchange configuration ($t=24$ s).}
   		\label{fig:exp2_config_exchange}
   \end{subfigure}
   \begin{subfigure}{0.32\linewidth}
   		\centering
    	\includegraphics[scale=1]{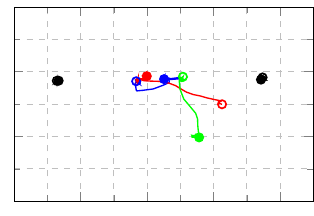}                             
   		\caption{Trajectories. Each square is $5$ m $\times$ $5$ m.}
   		\label{fig:exp2_trajectories}
   \end{subfigure}   
   \caption{Spatial configurations and routing solutions of the mobile infrastructure during the reconfiguration experiment. After the replacement agent enters the system, the network nodes go back to the initial configuration. Black circles represent task agents, while red, blue, and green (replacement agent) circles represent network agents. In the trajectories, solid dots represent initial positions, while hollow dots correspond to end positions. The lines connecting these dots show the trajectories taken by the agents during the experiment.}
   \label{fig:exp2_snapshots}
\end{figure*}

The resulting spatial and routing configurations can be seen in Fig. \ref{fig:exp3_snapshots}. As seen in Fig. \ref{fig:exp3_config_initial}\textendash\ref{fig:exp3_config_extended}, compared to the starting configuration, once the task agents pull away, the routing probabilities favor a flow of information towards the network agents, avoiding direct communication between the task agents. The trajectories taken during the experiment are shown in Fig. \ref{fig:exp3_trajectories}. Overall, the network agents assume a somewhat triangular inner circle formation, tracking the rotation of the task agents. At any given instance, the execution of the probabilistic robust routing solution requires a sampling of probabilities. Though occurring in the same way, this is something that we have not explicitly shown in our previous experiments. Fig. \ref{fig:exp3_routing_executed} shows the exact sampling of the routing configuration Fig. \ref{fig:exp3_config_extended} at a specific time instance  ($t=145$ s), which is then employed at the communication interface. As previously discussed, the probability of direct communication between task agents is very low, and thus traffic moves through the inner network agents before reaching the destination task agents. 

The average data rate across the network observed during the experiment is shown in Fig. \ref{fig:exp3_rate}. As the experiment begins, the task agents move away to their specified patrol radius. As they do this, the data rate drops, at around $t=25$~s, they begin the patrol and at the same time, the mobile infrastructure team begins operating. As the network team repositions, the data rate starts recovering to a baseline level of $5$~Mbps at time $t=60$~s, which is kept stable during the whole duration of the experiment. Similar behavior occurs with the delay experienced in the network, shown in Fig. \ref{fig:exp3_delay}. In this case, as the task agents pull away from the center, the delay increases and is brought back to stable levels under $115$~ms when the mobile infrastructure team begins operating. Hence, the MID system is adaptive to complex trajectories of multiple task agents, ensuring consistent data rates and delays.

\subsection{Robust autonomous reconfiguration}

Finally, we conduct an experimental test to study network robustness and reconfiguration. This experiment is in the same vein as the simulation described in Section \ref{sec:sim2_replacement}. The experiment setup is as follows. Two stationary task agents (10.42.0.1 and 10.42.0.2) are placed at a separation of about $30$~m. They take off and keep hovering without moving from their initial positions. Two network agents (10.42.0.3 and 10.42.0.4) are then deployed using our MID system to support traffic between the task agents. An additional network agent (10.42.0.5) is placed aside as a backup. At a given point during the experiment, one of the network agents (10.42.0.3) deactivates itself and the backup UAV is activated. The mobile infrastructure reconfigures itself dynamically and without human input  to support the communication needs of the task agents.

The trajectories, routing, and spatial configurations during the experiment are shown in Fig. \ref{fig:exp2_snapshots}. In the initial configuration, the network agents position themselves in between near the middle of the two task agents. The optimal positioning is such as to provide a certain level of redundancy, instead of acting purely as a two-hop relay between the task agents. As such, the optimal routing configuration  splits approximately half of the traffic from each task agent between the two network agents. When the network agent 10.42.0.4 drops out and the substitute network agent 10.42.0.5 is activated into the MID system, the network nodes reposition themselves. Agent 10.42.0.3 progressively moves to the position of the network agent that left the network, and the new network agent takes the previous position of agent 10.42.0.4. Once the repositioning stabilizes, the final configuration is similar to the initial configuration.

\begin{figure}[t]
	\centering
    \includegraphics[scale=1]{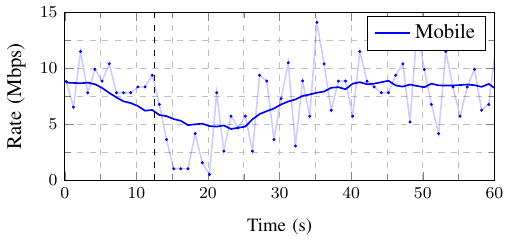}     
	\caption{Achieved data rate for the reconfiguration experiment. The instantaneous rate collected at  intervals of one second and a moving average over $20$ seconds are shown. At  $t=13$~s the network agent 10.42.0.4 drops from the system and is subsequently substituted by agent 10.42.0.5. The MID team adapts during the exchange to hold an average communication rate slightly above $4.5$~Mbps and stabilizes eventually to an average rate of $8$~Mbps, which is comparable to the level before the agent dropout.}
    \label{fig:exp2_rate}
\end{figure}

The data rate attained during the experiment can be seen in Fig. \ref{fig:exp2_rate}. When the starting network agent deactivates itself ($t=13$~s), the average data rate drops and then progressively increases as the new network agent enters the system and the mobile infrastructure reconfigures itself. This effect is more evident when looking at the instantaneous data rate. When the network agent leaves the system ($t=13$~s), the data rate immediately drops drastically. As the network reconfigures itself, the data rate progressively recovers, with the average communication rate remaining above $4.5$~Mbps during the exchange and stabilizing to an average rate of $8$~Mbps, comparable to the pre-exchange level.

Through the execution of this experiment, we have demonstrated how the MID system reacts when faced with the deactivation of a network agent, effectively recovering the network from system failure. As in the previous two experiments, our experimental finding not only corroborates the behavior observed in simulations, but also provides evidence for the practical viability of the proposed mobile infrastructure on demand system.

\section{Conclusions}

We have introduced the MID system, utilizing UAVs to deliver wireless connectivity on demand. The UAVs serve as communication relays, forming an ad-hoc network and dynamically adapting their routing decisions and positions to guarantee the required quality of service demanded by the system users. Simulated and experimentally validated, the MID system has demonstrated its capability to serve many scenarios. Distilled into three distinctive features: \textbf{(F1)} expanding the operational range, \textbf{(F2)} supporting tasks involving complex mobility patterns, and \textbf{(F3)} exhibiting robustness via autonomous reconfiguration. Simulations and experiments have shown that MID outperforms fixed infrastructure solutions, offering greater efficiency with fewer infrastructure elements, enhanced area coverage, increased data rates, lower delays, and robustness to node failure. Given these observations, the MID system holds considerable potential in enhancing the quality of service and bridging connectivity gaps in previously underserved areas.

\bibliographystyle{IEEEtran}
\bibliography{bib}

\appendix

\section{Appendix}
\subsection{Optimal routing as an SOCP}\label{sec:appendixNP}
The following problem is expressed in  the standard convex SOCP formulation \cite[Chapter 4.4.2]{boyd2004convex} which can be solved using off-the-shelf optimization solvers, including the \texttt{cvxpy} library used in our ROS implementation.
\begin{equation}\label{eq:minfz}
\begin{aligned}
	\underset{\mathbf{z}}{\text{minimize}} & &&  \mathbf f^T \mathbf z \\
	\text{subject to}&&&  \ \|{\mathbf C_p\mathbf z+\mathbf r_p}\| \leq \mathbf d_p^T\mathbf z + n_p, \quad p = 1,\dots, P 		
\end{aligned}
\end{equation}
To bring the optimal routing problem \eqref{eq:maxs} into this standard form, let us define $P=KL+2L+2KL^2+1$, with $K$ being the number of data flows and $L=N+M$ the total number of agents. Also define $\mathbf z=(\bm \alpha^T,s)^T$ with the probabilities $\alpha_{ij}^{k}$ collected in vector $\bm \alpha$. Then, the costs of \eqref{eq:minfz} and \eqref{eq:maxs} become identical by setting $\mathbf f=(\mathbf 0^T,-1)^T$.  
If $\mathbf C_p$ and $\mathbf r_p$ are set to zero, then the corresponding constraint in \eqref{eq:minfz} becomes linear in $\mathbf z$, and thus also linear in $\bm \alpha$. Hence, \eqref{eq:minfz} admits the equivalent form  

\begin{align}\label{eq:SOCP}
	\underset{\bm \alpha,s\geq 0}{\text{maximize}} & &&  s \\
	\text{subject to}&&&  \|\mathbf C_{i}^k \bm \alpha\|\leq \bm \alpha^T \mathbf d_{i}^{k} -s -m_{i}^k \quad \forall i,k\nonumber \\ 	
				&&& \mathbf M\bm \alpha\leq\mathbf  1\nonumber \\
				&&& \mathbf 0\leq \bm \alpha\leq \mathbf 1\nonumber
\end{align}

by redefining $\mathbf d_p=( (\mathbf d_{i}^k)^T,-1)^T$, $n_p=-m_i^k$, and $\mathbf C_{i}^{k} \bm \alpha=\mathbf C_p\mathbf z$ (removing the last column of $\mathbf C_p$) for the first $K(M+N)$ constraints, and then transforming the remaining ones into linear constraints and grouping them as $s\geq 0$, $\mathbf M\bm \alpha\leq\mathbf n$ and $\mathbf 0\leq \bm \alpha\leq \mathbf 1$.  With \eqref{eq:minfz} and \eqref{eq:SOCP} being equivalent, we consider henceforth  problem \eqref{eq:SOCP} as our definition of an SOCP. 

Next, we show how to set the constraints of  \eqref{eq:maxs} in the form of \eqref{eq:SOCP}. In particular, \eqref{eq:maxs} has the constraint
\begin{align}\label{eq:recall_maxs}
\bar b_i^k(\boldsymbol{\alpha},\mathbf{x}) - m_i^k-s\geq  \zeta_i^k \sqrt{\tilde b_i^k(\boldsymbol{\alpha},\mathbf{x})}
 \end{align}
 with $\zeta_i^k \triangleq \Phi ^{-1} (\epsilon_i^k)$ being the percentile of the error. Notice that \eqref{eq:recall_maxs} includes $\bar b_i^k(\boldsymbol{\alpha},\mathbf{x})$ which is linear in $\bm \alpha$, so it can be rewritten as  $\bar b_i^k(\boldsymbol{\alpha},\mathbf{x})= \bm \alpha^T \mathbf d_{i}^{k}$ by properly defining $\mathbf d_{i}^{k}$ in terms of the average rate $\bar R_{ij}$ (see Algorithm \ref{alg:SOCPplanner}).
Similarly, $\tilde b_i^k(\boldsymbol{\alpha},\mathbf{x})$ in \eqref{eq:tildeb} is quadratic in $\bm\alpha$ so that $\zeta_i^k \sqrt{\tilde b_i^k(\boldsymbol{\alpha},\mathbf{x})}$ can be rewritten using the $\ell_2$-norm in the form
\begin{align}
\zeta_i^k \sqrt{\tilde b_i^k(\boldsymbol{\alpha},\mathbf{x})} = \|\mathbf C_{i}^{k}  \bm \alpha\|
\end{align}
by properly defining the diagonal matrix $\mathbf{C}_{i}^{k}$ in terms of the rate variances $\tilde R_{ij}$ and the percentile $\zeta_i^k$ (also in Algorithm \ref{alg:SOCPplanner}). Under these definitions, it is apparent that the constraints in \eqref{eq:recall_maxs} and \eqref{eq:SOCP} coincide.The remaining  $\boldsymbol{\alpha} \in \mathcal{A}$ and $s \geq 0$ in \eqref{eq:minfz} are linear, and thus they can  also be written as in \eqref{eq:SOCP}. More details on how to select the matrix $\mathbf M$ and vector $\mathbf n$ for that purpose are given in Algorithm \ref{alg:SOCPplanner}, where $\mathbf{e}_i$ are vectors of the canonical basis.

\subsection{Optimal connectivity as an SDP}\label{sec:appendixCP}

Let us start by arguing that the connectivity of a graph is strengthened by maximizing the second eigenvalue of the graph Laplacian, and then reformulate this maximization problem to solve it with a standard convex optimization algorithm. By construction, the smallest eigenvalue of the Laplacian is zero, and its corresponding eigenvector is the vector $\mathbf 1$ which has  all elements equal to one. Furthermore, the multiplicity of the null eigenvalue gives the number of disconnected sub-graphs. Indeed, if the graph has $R$ disconnected regions, the Laplacian will be block diagonal with each of the $R$ blocks being the Laplacian of each sub-graph.
For instance,  if a network with $5$ nodes is such that nodes  $1$, $2$, and $3$ are pairwise connected but isolated from the pair of nodes $4$ and $5$, then the eigenvalue corresponding to the eigenvector $\mathbf{v}=(1, 1,1,0,0)$ will also be null. This is because the first three ones will nullify the $3\times 3$ Laplacian of the sub-graph that contains nodes $1, 2,$ and $3$, and will not multiply the other block. Thus, the eigenvalue associated with $\mathbf{v}=(1, 1,1,0,0)$ is also null. In this case, the null eigenvalue has multiplicity $R=2$ with eigenvalues $\mathbf{v}=(1,1,1,1,1)$ and $\mathbf{v}=(1,1,1,0,0)$. 
For the network to be fully connected, the null eigenvalue must have multiplicity $1$. Using a continuity argument, a second smallest eigenvalue being close to zero would unveil a fully connected graph, with some weak links, since the network will disconnect as this eigenvalue tends to zero. Finally, the magnitude of the second eigenvalue indicates the strength of the network.

\begin{algorithm}[t]
\caption{\emph{Network Planner (SOCP Form)}}\label{alg:SOCPplanner}
\begin{algorithmic}
\State  Input $\bar R_{ij},\ \tilde R_{ij},\ m_i^k,\ \epsilon_i^k$
\For{$k=1,\ldots,K$}
\For{$i=1,\ldots,L$}
\State Set $\textbf{c}_{i}^k=\textbf{d}_{i}^k=\textbf{0}$
\For{$j=1,\ldots,L$}
\State Set $m=(k-1)L^2+(i-1)L+j$
\State Set $n=(k-1)L^2+(j-1)L+i$
\State Set $p=i$ 
\State Set $q=L+j$
\State $\mathbf c_{i}^k=\mathbf c_{i}^k+\textbf{e}_m\tilde  R_{ij}+\textbf{e}_n \tilde  R_{ji}$
\State $\mathbf d_{i}^k=\mathbf d_{i}^k+\textbf{e}_m\bar  R_{ij}-\textbf{e}_n\bar  R_{ji}$
\State $\mathbf M=\mathbf M+\textbf{e}_{p}\textbf{e}_m^T+\textbf{e}_{q}\textbf{e}_m^T$
\EndFor
\State Set $\mathbf C_{i}^k=\diag(\mathbf c_{i}^k)/(\Phi^{-1}(\epsilon_i^k))$ 
\EndFor
\EndFor
\State Solve SOCP \eqref{eq:SOCP} and output $\bm \alpha=\textrm{vec}(\alpha_{ij}^k)$ 
\end{algorithmic}
\end{algorithm} 

In the direction of maximizing the second eigenvalue, consider  the family of semidefinite programs (SDPs). These are convex problems that admit the following standard form  \cite[Chapter 4.6.2]{boyd2004convex},
\begin{equation}\label{eq:minfzSDP}
\begin{aligned}
	\underset{\mathbf{z}}{\text{minimize}} & && \mathbf f^T \mathbf{z} \\
	\text{subject to}&&& \mathbf  F_0+z_1\mathbf F_1+z_2\mathbf F_2+\ldots+z_{P}\mathbf F_P \preceq 0 \\ 	
				&&& \mathbf M\mathbf z=\mathbf r
\end{aligned}
\end{equation}%
Both the semidefinite and the linear constraints can be separated into multiple constraints by selecting matrices $\mathbf F_p$ and $\mathbf M$ as block diagonal matrices, and the variables can be included, or not, in each of these multiple constraints by setting to zero, or not, the corresponding blocks. Even a nonnegative constraint can be incorporated using a block of dimension one. Hence we are going to adopt the following equivalent problem as our definition of an SDP  
\begin{equation}\label{eq:strangeSDP}
\begin{aligned}
    	\underset{\gamma,\mathbf l,\mathbf a,\mathbf x,\mathbf u\geq 0,\mathbf v\geq 0}{\text{maximize}} & &&  \gamma \\
	\text{subject to}&&&  \gamma \mathbf I-\sum_{n=1}^{L^2} l_n \mathbf F_n\preceq 0 \\ 	
				&&& \mathbf x-\mathbf{\hat x}+\mathbf u= \Delta \mathbf 1 \\
				&&&\mathbf{\hat x}- \mathbf x+\mathbf v=\Delta \mathbf 1 \\
				&&& \mathbf l=(\mathbf C-\mathbf I)\mathbf  a \\
				&&& \mathbf a=\mathbf B\mathbf x+\mathbf d 		
\end{aligned}
\end{equation}
where we also separated $\mathbf z=(\gamma,\mathbf l^T,\mathbf a^T,\mathbf x^T,\mathbf u^T,\mathbf v^T)^T$ and renamed the blocks of the matrices $\mathbf F_p$ and $\mathbf M$ and vector $\mathbf r$ in a way that will be useful to prove the next equivalence with the connectivity problem \eqref{eq:fiedlermax}. Finally, for the  costs of \eqref{eq:strangeSDP} and  \eqref{eq:minfzSDP} to coincide. we must select $\mathbf f=(-1,\mathbf 0^T)^T$.

To put \eqref{eq:fiedlermax} in the form of \eqref{eq:strangeSDP}, define $\mathbf l=\textrm{vec} (\hat{ \mathbf L})$ and $\mathbf a=\textrm{vec} (\hat{\mathbf A})$, where the $\mathbf y=\textrm{vec}(\mathbf Y)$ operator returns a vector $\mathbf y$ formed by the concatenation of the columns of its matrix argument $\mathbf Y$.
The cost of \eqref{eq:fiedlermax} is already equal to that of \eqref{eq:strangeSDP}. The constraint  $\mathbf P^T\mathbf L \mathbf P \succeq \gamma \mathbf I$ is equivalent to $\gamma \mathbf I- \mathbf P^T \mathbf L \mathbf P\preceq \mathbf 0$ that can be rewritten as in \eqref{eq:strangeSDP} by properly defining the matrices $\mathbf F_n$ in terms of the columns of $\mathbf P$ as it is detailed in Algorithm \ref{alg:SDPplanner}. The constraint $\|\mathbf x-\hat{\mathbf x}\|_\infty\leq \Delta$ can be  put as  the system of linear equations in \eqref{eq:strangeSDP} by adding the nonnegative auxiliary variables  $\mathbf u$ and $\mathbf v$.
The remaining  constraints in \eqref{eq:fiedlermax} are linear in $\mathbf a$, $\mathbf l$, and $\mathbf x$, so that they can be rewritten as $\mathbf l=(\mathbf C-\mathbf I)\mathbf  a$ and  $\mathbf a=\mathbf B\mathbf x+\mathbf d$ by properly defining $\mathbf B$,  $\mathbf C$, and $\mathbf d$. 
More details, including the construction of  $\mathbf B$,  $\mathbf C$, and $\mathbf d$ can be found in Algorithm \ref{alg:SDPplanner}, where the symbol $\otimes$ stands for the Kronecker product.

\begin{algorithm}[t]
\caption{\emph{Connectivity Planner (SDP Form)}}\label{alg:SDPplanner}
\begin{algorithmic}
\State  Input $\hat{R}_{ij}$, $\nabla_{x_i} \hat{R}_{ij}$, $\nabla_{x_j} \hat{R}_{ij}$, $\mathbf{\hat x},$ and $\mathbf P$ with rows $\mathbf p_i$

\State Set $  \mathbf B=\mathbf 0,\ \mathbf C=\mathbf 0$, $\mathbf d=\mathbf 0$ 
\For{$i=1,\ldots,L$}
\State $m=(i-1)L+i$
    \For{$j=1,\ldots,L$}
        \State $n=(i-1)L+j$
        
        \State  $\mathbf F_{n}=\mathbf p_i \mathbf p_j^T$
       \State $\mathbf C=\mathbf C+\mathbf e_{m}\mathbf e_n^T$
        \State Set $\mathbf g_{n}=\nabla_{x_i} \hat{R}_{ij}$  and $\mathbf h_{n}=\nabla_{x_j} \hat{R}_{ij}$
        \State  $\mathbf B=\mathbf B+\mathbf e_n \left((\mathbf e_i\otimes \mathbf g_n)^T+(\mathbf e_j\otimes \mathbf h_n)^T\right)$ 
        \State $\mathbf d=\mathbf d+ \mathbf e_n(\hat R_{ij}-\mathbf g_n^T\hat {\mathbf x}_{i}-\mathbf h_n^T\hat{ \mathbf x}_j)$
\EndFor
\EndFor
\State Solve SDP \eqref{eq:strangeSDP} and output $\bf x$
\end{algorithmic}
\end{algorithm} 

\subsection{Mapping to a common reference}
\label{sec:appendixCommonReference}

 By default, the position $x_i$ of a UAV is referenced to its own local reference system $s_i$ where they were initialized. In our experiments, we define a nearby common reference $s_r$ to be shared by all UAVs. We select this point as a clear visual reference within the zone where we are going to carry out the experiments. In order to obtain a transformation $H_i^r$ that maps the position $x_i$ to this common reference system $s_r$, the mapping node follows the following procedure. It takes the GPS coordinates of $s_r$ as input parameters and receives from PX4 the GPS coordinates of the current position $s_p$. With these parameters, it computes the transformation $H_g^r$  between the global GPS reference $s_g$ and the common reference $s_r$. It also computes the transformation $H_g^p$ between the global GPS reference $s_g$ and the current GPS coordinates $s_p$. Then it obtains from PX4 the transformation $H_i^p$ from the initialization to the current position. These three transformations give the sufficient information to compute $H_r^i=(H_g^r)^{-1}H_g^p(H_i^p)^{-1}$. Hence, using $H_r^i$ it is possible to map the position $x_i$ to the common reference $s_r$, transforming the position of the UAV, from its internal reference system $s_i$, to the common reference system $s_r$. At a high level, all variables are expressed in these common coordinates when processed by the network and connectivity planners in ROS, and the mapping node is called to communicate to and from the lower-level flight controller.

\end{document}